\documentclass[10pt,twocolumn]{article}

\usepackage[
  a4paper,
  top=2.0cm,
  bottom=2.0cm,
  left=1.8cm,
  right=1.8cm,
  columnsep=0.7cm
]{geometry}
\usepackage[T1]{fontenc}
\usepackage[utf8]{inputenc}
\usepackage{lmodern}
\usepackage{microtype}
\usepackage{authblk}

\usepackage{amsmath,amssymb,amsfonts}
\usepackage{graphicx}
\graphicspath{{figures/}{deliverables/v6_kg_analysis/}{deliverables/v5_kg_analysis/}{deliverables/v4_kg_analysis/}{deliverables/v3_kg_analysis/}}
\usepackage{algorithm}
\usepackage{algorithmic}
\usepackage{textcomp}
\usepackage{xcolor}
\usepackage{booktabs}
\usepackage{multirow}
\usepackage{pifont}
\usepackage{array}
\usepackage{tabularx}
\usepackage{placeins}
\usepackage{stfloats}
\usepackage{balance}
\usepackage{tikz}
\usetikzlibrary{arrows.meta,positioning,shapes.geometric,fit,calc,backgrounds}
\usepackage{comment}

\usepackage[authoryear,round]{natbib}
\let\cite\citep
\usepackage{hyperref}
\usepackage[nameinlink,noabbrev]{cleveref}
\definecolor{arxivlink}{RGB}{0,70,140}
\hypersetup{
  colorlinks=true,
  linkcolor=arxivlink,
  citecolor=arxivlink,
  urlcolor=arxivlink,
  pdftitle={TRACE-CTI: Auditable Post-Extraction Governance of TTP Claims with Knowledge Graphs},
  pdfauthor={Federico Valletta, Giacomo Longo, Enrico Russo, Alessio Merlo}
}

\def\cmark{\ding{51}}
\def\xmark{\ding{55}}
\def\pmark{\textbf{$\sim$}}

\newcommand{\tool}{TRACE-CTI}

\title{\textbf{\tool{}: Auditable Post-Extraction Governance of TTP Claims with Knowledge Graphs}}

\author[1]{Federico Valletta}
\author[1]{Giacomo Longo}
\author[1]{Enrico Russo}
\author[1]{Alessio Merlo\thanks{Corresponding author: \href{mailto:alessio.merlo@unicasd.it}{alessio.merlo@unicasd.it}}}
\affil[1]{CASD -- School of Advanced Defense Studies, Piazza della Rovere 83, 00165 Rome, Italy}
\date{}

\begin{document}

\twocolumn[
\begin{@twocolumnfalse}
\maketitle

\begin{abstract}
Security Operations Centers increasingly rely on automated mapping of Cyber Threat Intelligence reports to MITRE ATT\&CK, yet extractor outputs remain fallible and are often stored without the evidence, provenance, and validation history needed to decide whether an individual mapping should be trusted. We present \tool{}, a post-extraction claim-governance framework that preserves run-level Predictions, aggregates them into configuration-level GraphAssertions, materializes setup-deduplicated corroboration as ConsensusAssertions, and exposes only GraphAssertions backed by policy-compliant validation grounds. The framework retains native evidence granularity, complete extraction provenance, versioned trust decisions, and non-destructive revocation history. We evaluate \tool{} on two public CTI corpora comprising 65 reports and 5,303 sentences, using a controlled $2\times3$ matrix of retrievers and generator families incrementally ingested across six GraphVersions. All setups are incorporated without schema modification; provenance paths remain complete, operational scopes remain disjoint, and every trusted GraphAssertion has an active qualifying validation ground. Cross-generator-family setup pairs exhibit greater output diversity than same-family pairs. At the final graph state, increasing setup support from $k\geq1$ to six-setup unanimity raises gold-aligned precision from 25.3\% to 90.6\%, while recall decreases from 88.2\% to 16.3\%. The graph also directly answers seven questions about provenance, trust, versioning, dependency, disagreement, and review-queue that the evaluated minimal flat output cannot fully answer without enrichment or reprocessing. These results support explicit, auditable governance of extracted TTP claims; the observed corroboration trajectory is descriptive and does not establish statistical independence or a causal model-family effect.
\end{abstract}


\vspace{0.8cm}
\end{@twocolumnfalse}
]

\section{Introduction}
\label{sec:intro}

Cyber Threat Intelligence (CTI) reports describe how real-world adversaries operate, including the tools they use, the actions they perform, and the systems they target. Security Operations Centers (SOCs) use this information to support activities such as threat hunting, detection engineering, and incident prioritization. However, CTI reports are typically written in unstructured prose, whereas operational security tools require structured data.

A common way to bridge this gap is to map passages from CTI reports to the tactics and techniques defined by MITRE ATT\&CK~\cite{strom2020mitre}. Producing these mappings manually is time-consuming and requires specialized expertise~\cite{orbinato2022automatic}. This has motivated a broad range of automated TTP-extraction methods, from rule-based systems to supervised classifiers and, more recently, Large Language Models (LLMs)~\cite{husari2017ttpdrill,satvat2021extractor,orbinato2022automatic,buchel2025sok}. Despite this progress, automated extraction remains imperfect.

A recent systematic evaluation reports that performance across more than forty methods plateaus at approximately 70\% F1, with ambiguity and limited labeled data remaining major sources of error~\cite{buchel2025sok}. LLM-based extractors introduce an additional concern: they may produce plausible ATT\&CK labels that are not adequately supported by the source text~\cite{ji2023hallucination}. Retrieval-augmented generation can improve the context available to a model, but it does not guarantee that the generated output is correct or grounded in the retrieved evidence~\cite{lewis2020rag}. This creates a problem that extraction accuracy alone does not address.

Once an extractor emits an ATT\&CK label, a SOC must decide whether that output is merely a model prediction or can be treated as trusted intelligence. An incorrect or insufficiently validated mapping may propagate into downstream operational processes as if it were reliable knowledge. This is particularly problematic in environments where false positives already contribute to alert overload and analyst fatigue~\cite{alahmadi2022falsepositives,sundaramurthy2015burnout}.

Deciding whether an extracted mapping can be trusted, therefore, requires more than a confidence score. An analyst must be able to determine which passage supports it, which model and configuration produced it, whether other distinct or heterogeneous extraction setups agree with it, and which downstream knowledge would be affected if a source or setup were later found to be unreliable. These questions concern the \textsl{lifecycle and governance} of extracted knowledge rather than the accuracy of the extractor itself.

Existing extraction pipelines are not designed around this distinction. They commonly store outputs as flat sentence--label records or directly incorporate extracted entities and relations into a knowledge base. In the first case, the information required to audit, compare, version, and revoke individual mappings is often incomplete. In the second case, a fallible model output may become indistinguishable from a validated fact. Consequently, improving precision, recall, or F1 does not by itself solve the operational problem: there is still no explicit and auditable boundary between what a model predicted and what the system is prepared to trust. Therefore, the unresolved problem addressed in this work is the following:

\begin{quote}
\emph{How can fallible outputs from heterogeneous TTP extractors be transformed into structured CTI knowledge without losing their evidence and provenance, and without silently treating predictions as trusted facts?}
\end{quote}

A solution to this problem should satisfy four requirements:
\begin{enumerate}
\item every run-level output should remain traceable to the exact source text and extraction process that produced it;
\item raw predictions and trusted knowledge should remain explicitly separated;
\item the basis on which a model-derived assertion becomes trusted should be recorded and inspectable;
\item the knowledge base should remain auditable as new reports, models, extraction results, validation decisions, and revocations are introduced.
\end{enumerate}

In this paper, we introduce \textbf{\tool{}} (\underline{T}rust-aware, \underline{R}evocable, and \underline{A}uditable \underline{C}laims with \underline{E}vidence for \underline{CTI}) to fulfill these requirements and address the post-extraction governance gap. \tool{} is not a new TTP extractor and does not depend on a specific language model. It receives the outputs of existing extractors and preserves each run-level output as an immutable \emph{Prediction} in a versioned knowledge graph. Each Prediction remains linked to its supporting evidence, extraction configuration, model run, retrieved context, and graph version. Predictions sharing the same normalized evidence unit, ATT\&CK target, extraction setup, and prompting method are aggregated into configuration-level \emph{GraphAssertions}. Agreement on the same evidence--technique target across distinct extraction setups is materialized through a \emph{ConsensusAssertion}. This representation preserves individual model observations while distinguishing them from the assertions governed by the knowledge graph and from the cross-setup evidence used to validate those assertions.

A prediction-derived GraphAssertion remains outside the trusted view unless an explicit, recorded validation event satisfies the active policy. Such validation may be grounded in a trusted gold annotation, corroboration from distinct extraction setups, or, when available, analyst review. A ConsensusAssertion records cross-setup support but does not become a trusted fact merely because it exists. This preserves an explicit boundary between observed model behavior, corroboration evidence, and knowledge exposed to downstream applications.

The key intuition behind the corroboration mechanism is that agreement may become more informative as the supporting setups diversify.
Multiple configurations sharing the same model family may reproduce correlated errors, whereas models from different families may exhibit less similar output behavior. Nevertheless, distinct configuration labels do not establish statistical independence. \tool{} therefore treats each retriever--generator setup as a distinct witness, records the composition of the supporting witness set, characterizes output diversity across same- and cross-generator-family setup pairs, and examines how gold-aligned correctness changes along the cumulative multi-setup agreement trajectory. Repeated runs, random seeds, and the RAG and RAG+FSP prompting variants associated with the same parent setup do not create additional setup witnesses.

We evaluate this idea using a controlled $2\times3$ matrix comprising two retrievers and three generator families. The six resulting extraction setups are incrementally ingested into the same graph over two public CTI corpora, covering 65 reports and 5,303 sentences. This design serves two purposes: it verifies whether the same schema and trust policy can govern outputs from heterogeneous extractors, and it characterizes output diversity across generator families and examines how gold-aligned correctness changes along the cumulative multi-setup agreement trajectory.

The evaluation also examines whether the resulting graph supports operational tasks that a flat collection of extraction outputs cannot directly support, including evidence tracing, version comparison, setup revocation, disagreement analysis, and the construction of analyst-review queues.

Based on this problem formulation, we investigate the following research questions:

\begin{itemize}
\item \emph{(RQ1) Auditable evolution.} Can \tool{} incorporate successive extraction setups without changing its schema or breaking the provenance paths of previously stored Predictions and GraphAssertions?
\item \emph{(RQ2) Extractor-independent governance and witness diversity.} Can the same representation and trust policy govern heterogeneous retriever--generator setups, and what evidence does the observed witness diversity provide about the informativeness of multi-setup corroboration?
\item \emph{(RQ3) Controlled promotion to trusted knowledge.} Can \tool{} maintain a trusted view in which every model-derived GraphAssertion is backed by an explicit and recorded validation event?
\item \emph{(RQ4) Operational utility.} Which provenance, validation, versioning, revocation, and review tasks become directly answerable through the knowledge graph but remain unavailable, incomplete, or require reprocessing in a flat extraction store?
\end{itemize}

This work makes four contributions:

\begin{itemize}
\item \emph{A claim-governance model.} We define a lifecycle that distinguishes immutable run-level Predictions, configuration-level GraphAssertions, cross-setup ConsensusAssertions, and the trusted view. This lifecycle preserves model outputs while preventing their silent promotion to trusted CTI knowledge.
\item \emph{A versioned and extractor-independent knowledge-graph schema.} We preserve a complete path from each governed GraphAssertion to its source evidence and to all Predictions, assertion configurations, model runs, validation records, and graph versions that support it, while retaining previous graph states and enabling non-destructive revocation.
\item \emph{A multi-witness validation mechanism evaluated over a
controlled retriever--generator matrix.}
We formalize trust through recorded validation events and
setup-deduplicated witness support, and empirically examine whether
increasingly restrictive multi-setup agreement identifies a smaller but
more precise set of evidence--technique targets.
\item \emph{An evaluation of graph-native SOC operations.} We assess the ability of \tool{} to support evidence auditing, trust inspection, version comparison, setup revocation, disagreement attribution, and analyst-review prioritization without re-running the extractors.
\end{itemize}

The remainder of the paper is organized as follows. Section~\ref{sec:background} introduces CTI reports, ATT\&CK mapping, and automated TTP extraction. Section~\ref{sec:related} positions \tool{} with respect to extraction pipelines, CTI knowledge graphs, threat-sharing platforms, and general provenance models. Section~\ref{sec:methodology} defines the claim-governance methodology and multi-witness trust model. Section~\ref{sec:framework} presents the framework architecture and knowledge-graph schema. Section~\ref{sec:evaluation} describes the experimental instantiation and reports the evaluation results. Section~\ref{sec:discussion} discusses the research questions, implications, and threats to validity. Finally, Section~\ref{sec:conclusion} concludes the paper.

\section{Background}\label{sec:background}

This section introduces the concepts needed to understand the post-extraction problem that \tool{} addresses. It explains how CTI reports are mapped to MITRE ATT\&CK, how automated TTP extraction produces and evaluates such mappings, and why predictive performance alone is insufficient to establish whether an individual output can be trusted.

\subsection{From CTI Reports to Structured Threat Knowledge}

Cyber Threat Intelligence (CTI) describes cyber threats, the actors and tools involved, and the behaviors observed during attacks. A major source is the threat report published by security vendors, research organizations, or incident-response teams after investigating a campaign, malware family, or incident. These reports help Security Operations Centers (SOCs) guide threat hunting, detection engineering, alert interpretation, and incident prioritization. However, they are written primarily as narrative documents: relevant evidence may span several sentences, the same behavior may be expressed in different ways, and technical descriptions often depend on context.

Operational tools instead require structured information that can be indexed, compared, and queried. MITRE ATT\&CK provides a widely adopted vocabulary for this purpose~\cite{strom2020mitre}. It organizes adversary behavior into \emph{tactics}, which describe objectives, and \emph{techniques} and \emph{sub-techniques}, which describe how those objectives are pursued. Stable identifiers allow reports, analysts, and security tools to refer consistently to the same behavior.

Mapping a CTI report to ATT\&CK, therefore, means identifying the passages that describe adversary behavior and associating them with the corresponding identifiers. This turns prose into structured knowledge that a SOC can use to compare campaigns, search for behaviors, and relate observations to defensive measures. Expert analysts traditionally perform this mapping by reading the report, interpreting each behavior in context, distinguishing similar techniques, and verifying that the selected label is actually supported by the text. The process is consequently time-consuming, difficult to scale, and dependent on specialized expertise~\cite{orbinato2022automatic}.

\subsection{Automated TTP Extraction and RAG}

Automated TTP extraction seeks to identify the ATT\&CK techniques or sub-techniques described by a sentence, text span, paragraph, or document. Because a single passage may describe multiple behaviors, the task is commonly formulated as multi-label classification: given CTI text, an extractor returns zero or more ATT\&CK identifiers.

The task remains difficult. ATT\&CK contains hundreds of labels, while public corpora cover only a limited and uneven subset. Similar techniques may use overlapping language; the same technique may be described differently across reports and platforms. Relevant context may occur outside the sentence being classified, and CTI combines natural language with malware names, commands, paths, protocols, and other domain-specific elements.

Existing methods include rule- and ontology-based systems, such as TTPDrill~\cite{husari2017ttpdrill}; supervised classifiers, such as rcATT~\cite{legoy2020rcatt,orbinato2022automatic}; and generative Large Language Models (LLMs), which directly emit one or more ATT\&CK identifiers. Although these families differ in how they derive labels, all produce predicted mappings between text and techniques. A recent systematic evaluation~\cite{buchel2025sok} shows that no family fully solves the task: performance remains constrained by label ambiguity, class imbalance, inconsistent annotation practices, and scarce labeled CTI data. Extraction errors must therefore be treated as an expected outcome rather than a rare exception.

Some LLM-based extractors use retrieval-augmented generation (RAG)~\cite{lewis2020rag}. A \emph{retriever} selects contextual information, such as potentially relevant ATT\&CK descriptions, and a \emph{generator} reads that context together with the CTI passage and produces the labels. The two components are replaceable and may affect the output differently: the retriever controls what information is shown, whereas the generator interprets it.

Retrieval may improve the available context, but it does not guarantee a correct or grounded output. An LLM may still generate a plausible but unsupported label~\cite{ji2023hallucination}, and retrieved material does not necessarily constitute the evidence on which the model relied. Auditing, therefore, requires three elements to remain distinct: the source passage being analyzed, the context supplied to the model, and the label it produced. RAG explains part of how a prediction was generated; it does not determine whether that prediction should be trusted, how it compares with outputs from other setups, or how it should be handled if its source or extractor is later found unreliable.

Accordingly, \tool{} is not another TTP extractor. It accepts outputs from different extraction approaches and governs what happens after generation. The pipeline used in our evaluation is one concrete source of predictions, not a framework requirement.

\subsection{Evaluation Corpora and Evidence Granularity}

TTP extractors are commonly evaluated on CTI corpora annotated by human experts. These \emph{gold labels} provide a reference against which predictions are classified as true positives, false positives, or false negatives. Precision measures the proportion of predictions matching the reference, recall measures the proportion of reference labels recovered, and F1 summarizes their balance. These metrics are essential for comparing extractors, but they characterize aggregate predictive performance rather than the status of an individual output, i.e., they do not show which exact words support a label, which setup produced it, whether a distinct extractor agrees, whether it has been reviewed, or what downstream knowledge would be affected by its withdrawal.

We use two public corpora that were used in prior TTP-extraction evaluations~\cite{buchel2025sok}. TRAM~v2 associates CTI sentences with ATT\&CK techniques~\cite{ctid2023tram}, whereas AnnoCTR links specific text spans to entities, tactics, and techniques~\cite{lange2024annoctr}. The distinction is important for auditability. A span-level annotation identifies the precise evidence and can safely be associated with its enclosing sentence. A sentence-level annotation states only that the sentence supports a mapping and cannot safely be projected onto a more precise span that the annotator did not identify. \tool{} therefore preserves the native granularity of the evidence. Section~\ref{sec:granularity} formalizes this policy.

\subsection{From Extractor Performance to Claim Reliability}

An extractor output records what a system predicted, but it does not by itself establish that the mapping is correct. Even a high-precision system produces false positives, and aggregate metrics cannot identify in advance which individual outputs are wrong. In a SOC, an unsupported ATT\&CK mapping may misdirect analyst attention, distort the interpretation of a report, or propagate inaccurate information into downstream analyses, adding to environments already affected by false-positive overload and analyst fatigue~\cite{alahmadi2022falsepositives,sundaramurthy2015burnout}.

Model confidence is not sufficient evidence of correctness. Scores may be uncalibrated and are not necessarily comparable across architectures or configurations; a model can be highly confident in an incorrect output. Retaining the source sentence is also insufficient because it may not identify the exact evidence or record the retriever, generator, prompt, run, retrieved context, and subsequent validation history.

The central distinction is therefore between \emph{extractor performance} and \emph{claim reliability}. The former measures how well an extractor performs over a corpus through precision, recall, F1, and related metrics. The latter concerns the basis on which a specific mapping may be accepted for operational use: which passage supports it, which setup and run produced it, whether it has human backing or analyst review, whether distinct setups corroborate it, and whether it can be revoked without destroying its history.

These questions require the prediction, its evidence, its provenance, and the events affecting its trust status to remain explicitly represented after extraction. \tool{} addresses this post-extraction problem by treating ATT\&CK mappings as inspectable claims rather than immediately converting model outputs into facts. The next section positions this lifecycle-oriented approach with respect to prior work.

\section{Related Work}
\label{sec:related}

\tool{} lies between automated CTI extraction and the systems that represent, trace, and share structured threat knowledge. Prior work addresses important parts of this path, but generally treats extraction, knowledge representation, provenance, and dissemination as separate concerns. We organize the literature around these research lines and focus on how each treats a model-generated mapping after extraction.

\subsection{Automated CTI Extraction and RAG}

Early CTI-processing systems extracted indicators, entities, and adversary behavior from reports. iACE identified indicators of compromise in open-source CTI~\cite{liao2016acing}; TTPDrill extracted threat actions using natural-language processing and domain knowledge~\cite{husari2017ttpdrill}; EXTRACTOR reconstructed attack behavior from reports~\cite{satvat2021extractor}; and rcATT classified text according to MITRE ATT\&CK~\cite{legoy2020rcatt}. Later studies examined linguistic patterns, supervised models, and automated ATT\&CK mapping~\cite{orbinato2022automatic,alam2023looking,li2024automated}. A recent systematic evaluation~\cite{buchel2025sok} compares rule-based, classification-based, and generative approaches, highlighting persistent limitations in labeled data, technique coverage, and annotation ambiguity.

This literature establishes how to generate and evaluate ATT\&CK mappings. Its main object is the predicted text--label association, typically assessed through precision, recall, and F1. The subsequent lifecycle of an individual prediction is usually outside scope: the output may be stored with a sentence or converted into a structured object, without remaining a distinct entity whose evidence, producer, validation history, trust status, and version can all be inspected and revised.

Retrieval-augmented generation (RAG) adds external context to generation~\cite{lewis2020rag} and is increasingly used in TTP extraction~\cite{buchel2025sok}. Retrieval may improve the information available to the model but does not prevent unsupported generation~\cite{ji2023hallucination}. Nor does it establish whether retrieved context supports the emitted label or whether the resulting prediction should be trusted, superseded, or revoked. Thus, RAG contributes to extraction provenance, but not a post-extraction trust policy.

\tool{} is complementary to these methods: extractors determine which labels are proposed, while \tool{} preserves and governs those proposals after generation.

\subsection{Cybersecurity Knowledge Graphs}

Cybersecurity knowledge graphs connect heterogeneous entities such as adversary behaviors, malware, vulnerabilities, and affected platforms. BRON~\cite{hemberg2020linking} integrates ATT\&CK, CWE, CVE, and related sources; POIROT~\cite{milajerdi2019poirot} aligns observed activity with known attack behavior over provenance graphs; and AttacKG and Piplai et al.~\cite{li2022attackg,piplai2020creating} construct attack-oriented graphs from CTI reports. MALOnt~\cite{rastogi2020malont} provides an ontology for malware-related concepts and relations.

These systems demonstrate the value of graph-based integration and analysis, but their primary objective is cybersecurity knowledge or the observation of activity. When content is automatically extracted, the model output is not generally represented as a separate lifecycle object. Consequently, a graph relation may not reveal whether it was manually curated, produced by a single extractor, corroborated across multiple setups, or later reviewed. \tool{} retains this distinction: the claim connecting report evidence to an ATT\&CK concept remains separate from the domain fact it may eventually support.

\subsection{Threat Sharing and General Provenance}

MISP and STIX/TAXII support the representation and exchange of structured threat information across tools and organizations~\cite{wagner2016misp,oasis2021stix}. They provide mechanisms for identifying, relating, marking, and updating shared objects, but operate mainly after information has been selected for representation or dissemination. Deciding which of several raw extractor outputs to add to that trusted collection remains an upstream responsibility. A trusted view produced by \tool{} can therefore feed a sharing platform while unsupported predictions remain in the audit graph.

General knowledge-graph models~\cite{hogan2021kg} provide reusable mechanisms for representing structured information, while PROV-O ~\cite{lebo2013provo} represents entities, activities, agents, and derivation relations. These foundations can describe where information came from and which process produced it, but are intentionally domain-independent. They do not define sufficient evidence for an ATT\&CK mapping, distinguish trusted from untrusted CTI claims, compare heterogeneous extractors, or specify the effect of revoking an extractor. \tool{} specializes these principles through a CTI-specific claim lifecycle and trust policy.

\subsection{Positioning of \tool{}}

The reviewed areas are complementary rather than direct competitors: extraction methods generate mappings; cybersecurity graphs organize domain knowledge; sharing platforms disseminate selected intelligence; and provenance models describe derivation. Table~\ref{tab:prior_work} summarizes their treatment of the post-extraction stage. The comparison is descriptive: an approach is not deficient because governance lies outside its objective. The relevant distinction is whether raw predictions, evidence, validation, trust, versioning, and revocation are treated as a single lifecycle.

\begin{table*}[!ht]
\caption{Positioning of \tool{} across research lines related to post-extraction CTI governance.}
\label{tab:prior_work}
\centering
\small
\renewcommand{\arraystretch}{1.08}
\setlength{\tabcolsep}{3pt}
\newcolumntype{Y}{>{\raggedright\arraybackslash}X}
\begin{tabularx}{\textwidth}{@{}p{2.6cm}p{4.4cm}YY@{}}
\toprule
\textbf{Research line} & \textbf{Representative work} & \textbf{Primary objective} & \textbf{Post-extraction treatment} \\
\midrule
Automated TTP extraction
& \cite{liao2016acing,husari2017ttpdrill,satvat2021extractor,legoy2020rcatt,orbinato2022automatic,alam2023looking,li2024automated,buchel2025sok}
& Extract ATT\&CK mappings from CTI text.
& Outputs are evaluated mainly as predictions; claim-level trust, history, and revocation are outside the primary objective. \\
\addlinespace
CTI knowledge graphs
& \cite{hemberg2020linking,milajerdi2019poirot,li2022attackg,piplai2020creating,rastogi2020malont}
& Represent and connect cybersecurity entities and behaviors.
& Domain knowledge is central; the extractor output and its validation history are not generally a distinct lifecycle. \\
\addlinespace
Threat-information sharing
& \cite{wagner2016misp,oasis2021stix}
& Represent and exchange structured CTI.
Information is typically handled after upstream selection; competing raw predictions remain outside the scope. \\
\addlinespace
RAG and general provenance
& \cite{lewis2020rag,ji2023hallucination,hogan2021kg,lebo2013provo}
& Provide retrieved context or general derivation tracking.
& They record parts of production and provenance but do not define a CTI-specific validation and trust policy. \\
\addlinespace
\textbf{\tool{}}
& \textbf{This work.}
& \textbf{Govern extractor outputs as evidence-linked, versioned CTI claims.}
& \textbf{Predictions, evidence, provenance, validation events, trust status, and revocation remain in one auditable lifecycle.} \\
\bottomrule
\end{tabularx}
\end{table*}

The gap is therefore not the absence of extraction, graph, sharing, or provenance mechanisms in isolation, but their integration around fallible model outputs. Existing lines do not jointly preserve each prediction, connect it to precise evidence and complete extraction provenance, separate it from trusted knowledge, record the event that changes its status, and retain its history after updates or revocation.

\tool{} addresses this gap by representing automated mappings as first-class claims whose provenance and trust status evolve explicitly. This provides the conceptual bridge to the methodology in the next section, which formalizes the claim lifecycle and multi-witness validation process.

\section{Methodology}\label{sec:methodology}

The previous sections identified a gap between producing an automated ATT\&CK mapping and treating that mapping as trusted CTI knowledge. This section defines the methodology used by \tool{} to govern that transition independently of a particular extractor or graph technology. The methodology preserves every model output, separates run-level observations from governed assertions and cross-setup consensus summaries, records the evidence used to validate each assertion, and allows the active trusted view to change without erasing its history.

\subsection{Methodological Overview and Design Goals}\label{sec:approach}

Let an extractor process a textual unit from a CTI report and emit one or more ATT\&CK identifiers. Each output is fallible: extraction performance remains limited by ambiguous labels, uneven training data, and annotation differences~\cite{buchel2025sok}, while generative models may produce plausible but unsupported mappings~\cite{ji2023hallucination}. The methodology, therefore, treats an extractor output as an observation to be preserved and assessed, not as a fact to be silently admitted to the knowledge base.

Four design goals follow from this premise.

\begin{enumerate}

\item \emph{Preservation.} Every output must remain traceable to the source text, extraction configuration, execution, retrieved context, and graph state that produced it.

\item \emph{Separation.} Run-level Predictions, configuration-level GraphAssertions, cross-setup ConsensusAssertions, validation evidence, and trusted knowledge must remain distinguishable. Storing a Prediction must not imply accepting it.

\item \emph{Explicit validation.}
A model-derived GraphAssertion may enter the trusted view only through a recorded validation event that satisfies the active policy. Such an event may originate from a trusted human annotation, analyst review, or corroboration by a policy-compliant set of distinct extraction setups.

\item \emph{Auditable evolution.} New reports, configurations, policies, and validation decisions must extend the audit history without destroying previous observations. At the same time, the active trusted view must be allowed to expand or contract after validation, supersession, or revocation.

\end{enumerate}

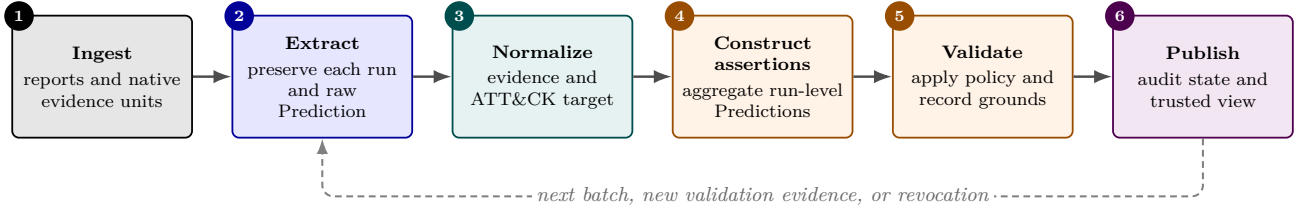
\begin{figure*}[t]
\centering
\resizebox{0.98\textwidth}{!}{%
\begin{tikzpicture}[
  font=\footnotesize,
  stage/.style={rounded corners=3pt, draw=#1!60!black, thick, fill=#1!10,
                minimum height=19mm, text width=25mm, align=center,
                inner xsep=1.5mm, inner ysep=2.5mm},
  num/.style={circle, fill=#1!60!black, text=white, font=\bfseries\scriptsize,
              minimum size=4.6mm, inner sep=0pt},
  flow/.style={-{Latex[length=2.8mm]}, line width=1.1pt, draw=black!70},
  back/.style={-{Latex[length=2.4mm]}, line width=0.9pt, draw=black!50,
               densely dashed, rounded corners=5pt},
  flbl/.style={font=\scriptsize\itshape, text=black!60, fill=white,
               inner sep=2pt, align=center}
]
\node[stage=black]                    (s1) {\textbf{Ingest}\\[2pt]reports and native\\evidence units};
\node[stage=blue, right=6mm of s1]    (s2) {\textbf{Extract}\\[2pt]preserve each run\\and raw Prediction};
\node[stage=teal, right=6mm of s2]    (s3) {\textbf{Normalize}\\[2pt]evidence and\\ATT\&CK target};
\node[stage=orange, right=6mm of s3]  (s4) {\textbf{Construct assertions}\\[2pt]aggregate run-level\\Predictions};
\node[stage=orange, right=6mm of s4]  (s5) {\textbf{Validate}\\[2pt]apply policy and\\record grounds};
\node[stage=violet, right=6mm of s5]  (s6) {\textbf{Publish}\\[2pt]audit state and\\trusted view};
\foreach \i/\c in {1/black,2/blue,3/teal,4/orange,5/orange,6/violet}
  \node[num=\c] at ([xshift=1.2mm,yshift=-0.2mm]s\i.north west) {\i};
\draw[flow] (s1) -- (s2);
\draw[flow] (s2) -- (s3);
\draw[flow] (s3) -- (s4);
\draw[flow] (s4) -- (s5);
\draw[flow] (s5) -- (s6);
\draw[back] (s6.south) -- ++(0,-9mm) -| (s2.south);
\node[flbl] at ($([yshift=-9mm]s2.south)!0.5!([yshift=-9mm]s6.south)$)
  {\small\itshape next batch, new validation evidence, or revocation};
\end{tikzpicture}}
\caption{Claim-governance lifecycle. The audit history is append-only, whereas the active trusted view is derived from the validation records and sources active at each version.}
\label{fig:method_blocks}
\end{figure*}

Figure~\ref{fig:method_blocks} summarizes the process as six governance stages.
In \emph{Ingest}, reports and their native evidence units are acquired.
In \emph{Extract}, each run-level output is preserved as a Prediction together with the provenance needed to reconstruct how it was produced.
In \emph{Normalize}, evidence references and ATT\&CK identifiers are mapped to a common representation so that outputs produced by different configurations can be compared consistently.
In \emph{Construct assertions}, Predictions referring to the same normalized claim are aggregated into configuration-level GraphAssertions, while agreement across distinct extraction setups is represented through ConsensusAssertions.
In \emph{Validate}, GraphAssertions are evaluated against the active policy, and the grounds supporting each trust decision are recorded.
Finally, in \emph{Publish}, the framework exposes both the complete audit state and the trusted view containing the GraphAssertions currently accepted for downstream use.

The following subsections progressively develop the methodological components that support this lifecycle.

\subsection{Claim Representation and State}
\label{sec:primitives}

The methodology is built around four connected elements: the textual \emph{evidence unit} supporting an ATT\&CK mapping; the \emph{provenance} of the extraction process that produced it{; the representation of the resulting \emph{claim} at three aggregation levels, namely the run-level \emph{Prediction}, the configuration-level \emph{GraphAssertion}, and the cross-setup \emph{ConsensusAssertion}; and the versioned \emph{graph state} in which the resulting objects are maintained. 
The three claim levels address complementary questions: \emph{what did a particular run output}, \emph{what did a particular extraction configuration} assert, and \emph{which normalized targets are supported by multiple setups}?

\paragraph{Evidence unit}\label{sec:terminology}
An evidence unit $e$ is the textual object on which an ATT\&CK mapping is asserted. It may be an entire sentence or a more precise span of evidence, depending on the native annotation or extraction granularity. The methodology never infers a finer trusted boundary than the source provides. A span may be linked to its enclosing sentence, but a sentence-level label is not converted into an artificial span.

\paragraph{Provenance}
The extraction provenance is described through the extraction setup, prompting method, and run.
An extraction setup $s$ is a versioned retriever--generator bundle together with the setup-level parameters shared across executions. A prompting method $m$ specifies how the retrieved context is presented to the generator. In the evaluated instantiation, $m\in\{\mathrm{RAG},\mathrm{RAG+FSP}\}$, where RAG+FSP adds five few-shot examples to the prompt. Their combination
\[
c := \langle s,m\rangle
\]
is the \emph{assertion configuration}: the aggregation level used to construct GraphAssertions. A run $r$ is one execution of configuration $c$ over a defined dataset and seed. Repeated runs or seeds belong to the same assertion configuration and therefore do not become additional GraphAssertions or additional setup witnesses.

\paragraph{Prediction}
A Prediction $p$ is the immutable record of one output produced by one run. After normalization, it contains or references the evidence unit $e$, ATT\&CK technique $a$, assertion configuration $c$, run $r$, model confidence when available, retrieved context, import batch, and graph version. A Prediction records model behavior and is untrusted by default.

\paragraph{GraphAssertion}
A GraphAssertion is the configuration-level claim
\[
g := \langle e,a,c\rangle,
\]
meaning that assertion configuration $c=\langle s,m\rangle$ associates evidence unit $e$ with ATT\&CK technique $a$. Predictions produced by repeated runs or seeds of the same configuration may therefore map to the same GraphAssertion. This aggregation preserves every run-level observation while preventing repeated executions from inflating the number of governed assertions.

\paragraph{ConsensusAssertion}
A canonical target is the configuration-independent pair
\[
u := \langle e,a\rangle.
\]
A ConsensusAssertion is a derived, materialized summary for a target supported by GraphAssertions associated with at least two distinct active extraction setups. It records the target, the supporting configuration-level GraphAssertions, the deduplicated setup support count, and the represented retriever and generator families. Its existence indicates cross-setup corroboration; it does not by itself imply membership in the trusted view. Stronger agreement levels, including strong consensus and unanimity, are read from the same materialized support information rather than represented as additional assertion types.

\paragraph{Graph state}
Let $b_t$ be the $t$-th import or governance batch and let $G_t$ be the corresponding logical graph state. $G_t$ contains the audit history accumulated up to $b_t$, together with the activation status of sources, configurations, GraphAssertions, ConsensusAssertions, and validation records at that version. A GraphVersion is not a destructive copy of the database but an identifiable state that can be reconstructed from the recorded history.

These definitions are storage-independent. Section~\ref{sec:framework} explains how \tool{} realizes them while retaining the schema and entity counts used in the experimental evaluation.

\subsection{Validation Events and Trust Policy}\label{sec:trust}\label{sec:normalisation}

Once Predictions have been organized into GraphAssertions and cross-setup support has been represented through ConsensusAssertions, the framework must determine which GraphAssertions may enter the trusted view and on what grounds. 
This task requires \emph{validation events}, which record the admission or withdrawal of trust; \emph{validation sources}, which identify the grounds supporting each decision; \emph{operational scopes}, which classify each GraphAssertion according to its current trust status; and the \emph{trusted view and policy invariant}, which formalize the conditions under which an assertion may be treated as trusted.

\paragraph{Validation events}
Trust is assigned to configuration-level GraphAssertions and exposed through the trusted view. 
Every reason for admitting or withdrawing an assertion is represented by a logically explicit, queryable validation record. Conceptually, each record represents a validation event, defined as
\[
v := \langle g,\tau,i,d,P,G_t,\sigma\rangle,
\]
where $g$ is the affected GraphAssertion, $\tau$ is the validation type, $i$ is the issuer, $d$ identifies the supporting grounds, $P$ is the policy under which the event was created, $G_t$ is the graph version, and $\sigma$ is its active or revoked status.

The methodology admits three validation sources:

\begin{itemize}

\item \emph{Gold validation}, based on a trusted human annotation supplied with a corpus;

\item \emph{Analyst validation}, based on an authorized review of the assertion and its evidence;

\item \emph{Multi-witness validation}, based on a ConsensusAssertion whose active cross-setup support satisfies the current corroboration policy.

\end{itemize}

These categories identify the provenance of trust. They do not make ConsensusAssertion a second trusted fact: the ConsensusAssertion supplies queryable corroboration evidence, while validated GraphAssertions remain the assertions exposed downstream.

\paragraph{Operational scopes}
Every GraphAssertion belongs to one of four exclusive operational scopes: \emph{gold}, \emph{validated}, \emph{prediction-only}, or \emph{deprecated}. Gold and validated assertions enter the trusted view. Prediction-only assertions have no qualifying validation record. Deprecated assertions remain in the audit graph but are excluded because their source, producer, or validation basis has been revoked. Within the validated scope, corroborated and unanimous subsets remain separately queryable; these are agreement-based decompositions, not additional top-level scopes.

Gold is assigned to a GraphAssertion when its normalized evidence--technique target is backed by a trusted human annotation. The annotation remains a separate gold instance and is not counted as a setup witness.

\paragraph{Trusted view and policy invariant}
Let $A_t$ denote the set of GraphAssertions active at version $t$, let $P_t$ be the policy active at that version, and let $V_t(g)$ be the validation records associated with GraphAssertion $g$. The trusted view is
\[
\begin{aligned}
T_t=\{g\in A_t\mid {}&\exists v\in V_t(g):\\
&\operatorname{active}(v,t)\land
\operatorname{satisfies}(v,P_t)\}.
\end{aligned}
\]

This definition yields the machine-checkable invariant used throughout the paper:

\begin{quote}
\emph{No prediction-derived GraphAssertion may appear in the trusted view unless it is backed by at least one active, recorded validation event satisfying the current policy.}
\end{quote}

The invariant prevents silent promotion while allowing a prediction-derived assertion to become trusted after explicit validation. If all qualifying validation records become inactive, the assertion leaves $T_t$ but remains available in $G_t$ for audit.

\begin{table}[t]
\caption{The four exclusive operational scopes of GraphAssertions. Agreement levels refine the validated scope but do not replace it.}
\label{tab:trust_scopes}
\centering
\small
\setlength{\tabcolsep}{2pt}
\begin{tabular}{@{}lp{4.0cm}c@{}}
\toprule
\textbf{Scope} & \textbf{Entry condition at version $t$} & \textbf{$\in T_t$} \\
\midrule
gold & backed by a trusted human-supplied annotation & yes \\
validated & backed by an active analyst or policy-compliant multi-witness event & yes \\
prediction-only & no active qualifying validation event & no \\
deprecated & source, configuration, or validation basis revoked & no \\
\bottomrule
\end{tabular}
\end{table}

Gold, analyst-reviewed, and multi-witness validation remain separately queryable. Likewise, \emph{corroborated}, \emph{strong-consensus}, and \emph{unanimous} describe measured agreement among active setups. They become grounds for trust only when the active policy requires the corresponding level of support.

\subsection{Multi-Witness Corroboration}\label{sec:multiwitness}\label{sec:trust-tiers}

Among the validation sources introduced above, multi-witness validation requires a precise definition of how support across distinct extraction setups is represented and assessed. 
The methodology therefore introduces the \emph{witness-support set}, which identifies the distinct active setups supporting a target; the \emph{support count and ConsensusAssertion}, which quantify and materialize that corroboration; \emph{witness diversity and policy requirements}, which determine whether the composition of the supporting setups satisfies the active policy; and the resulting \emph{agreement views}, which express increasingly restrictive levels of corroboration.

\paragraph{Witness-support set}
For a target $u=\langle e,a\rangle$, let $S_t$ be the set of active extraction setups at version $t$, and let $C(s)$ denote the assertion configurations whose parent setup is $s$. Its witness-support set is
\[
\begin{aligned}
W_t(u)=\{s\in S_t\mid {}&\exists c\in C(s),\\
&\exists g=\langle e,a,c\rangle:
\operatorname{active}(g,t)\}.
\end{aligned}
\]

The set is deduplicated by parent ExtractionSetup: RAG and RAG+FSP configurations associated with the same retriever--generator setup contribute at most one witness to $W_t(u)$, and repeated runs or seeds cannot add further votes. 

\paragraph{Support count and ConsensusAssertion}
The support count is 
\[
k_t(u)=|W_t(u)|.
\]
When $k_t(u)\geq2$, \tool{} materializes or versions one ConsensusAssertion for $u$. The ConsensusAssertion provides an efficient, queryable representation of the inclusive corroborated view and retains links to all supporting configuration-level GraphAssertions.

\paragraph{Witness diversity and policy requirements}
A distinct setup is a distinct witness identity, but distinctness does not imply statistical independence. Two setups may share a generator, retriever, prompt, training lineage, or source corpus and may therefore reproduce correlated errors. Ensemble methods motivate the use of diverse predictors~\cite{breiman2001random,dietterich2000ensemble}, but evidential independence cannot be assumed from configuration labels alone. The methodology, therefore, exposes both the size of support and the diversity of witnesses to the trust policy.

Let $F_G(W_t(u))$ and $F_R(W_t(u))$ denote the generator and retriever families represented in the support set. A policy may require, for example,
\[
k_t(u)\geq k_{\min},
\qquad |F_G(W_t(u))|\geq f_{\min},
\]
possibly together with analyst review, source restrictions, or other provenance constraints. The threshold $k_{\min}=2$ defines the first non-trivial corroborated view, but it is not treated as a universal correctness guarantee. The evaluation measures how precision and recall change as support and diversity requirements become stricter.

\paragraph{Agreement views}
For descriptive analysis, the framework retains three nested agreement views:
\[
\begin{aligned}
C_t^{(2)} &= \{u: k_t(u)\geq2\},\\
C_t^{(3)} &= \{u: k_t(u)\geq3\},\\
U_t &= \{u: W_t(u)=S_t\}.
\end{aligned}
\]
Thus $U_t\subseteq C_t^{(3)}\subseteq C_t^{(2)}$ whenever at least three setups are active. ConsensusAssertion nodes materialize $C_t^{(2)}$; stronger levels are derived from their support metadata. Membership in one of these views supplies validation evidence only when it satisfies $P_t$.

Disagreement is preserved rather than collapsed. If active configurations associate different techniques with the same evidence, each configuration-level GraphAssertion remains available, and distinct targets acquire separate support sets. Abstention contributes no support. This permits later analysis of whether disagreement is associated mainly with retrieval, generation, prompting method, or normalization.

\subsection{Versioned Evolution and Revocation}\label{sec:expansion}
Because witness support and trust decisions depend on which sources, setups, configurations, and validation records are active, the methodology must represent both how the graph evolves and how previously accepted support can be withdrawn. 
It therefore distinguishes \emph{versioned evolution}, which records successive graph states without overwriting prior history, from \emph{non-destructive revocation}, which updates active support and trust while preserving the records that explain earlier decisions.

More specifically, the methodology separates an append-only audit history from a non-monotonic active state. New Predictions, GraphAssertions, ConsensusAssertion states, validation records, and revocation records are appended; previous records are not overwritten. However, the active support set and trusted view may expand or contract when setups, datasets, configurations, or validation records are activated, superseded, or revoked.

Candidate setups may be organized as an expansion matrix. With retrievers $R$ and generators $L$, the cells of $R\times L$ define the setup identities used in the experimental instantiation. A prompting method selects an assertion configuration within a setup but does not create an additional setup witness. More generally, a configuration space may include a prompt family, a provider, a retrieval policy, a dataset source, or an analyst team. The matrix is an experimental device for controlling witness diversity; the governance model itself is agnostic to the number and ingestion order of setups.

A revocation is non-destructive. When a source, run, configuration, setup, or validation record is revoked, the framework records the revocation, updates the corresponding active states, recomputes affected support sets, and reevaluates policy-derived validation records. A GraphAssertion that no longer has an active qualifying record leaves the trusted view in the newly published version, while its previous status and complete provenance remain reconstructible.

The resulting property is therefore more precise than monotonic knowledge growth: \emph{the audit history grows monotonically, whereas the active trusted view is version-dependent}.

\subsection{Governance Procedure}
\label{sec:governance-procedure}

\begin{algorithm}[t]
\caption{Versioned claim-governance procedure.}
\label{alg:expansion}
\begin{algorithmic}[1]

\STATE \textbf{Input:} initial audit state $G_0$; ordered batches $B=(b_1,b_2,\ldots)$
\label{alg:expansion:input}

\STATE \textbf{Output:} versioned audit states $G_t$ and trusted views $T_t$
\label{alg:expansion:output}

\FOR{each batch $b_t \in B$}
\label{alg:expansion:foreach}

  \STATE $(E_t,X_t,H_t,R_t,P_t)
  \gets \mathrm{UnpackBatch}(b_t)$
  \label{alg:expansion:unpack}

  \STATE $O_t
  \gets \mathrm{PreservePredictions}(E_t,X_t)$
  \label{alg:expansion:predictions}

  \STATE $\widehat{O}_t
  \gets \mathrm{Normalize}(O_t)$
  \label{alg:expansion:normalize}

  \STATE $G_t'
  \gets \mathrm{ApplyStateChanges} (G_{t-1},R_t)$
  \label{alg:expansion:state}

  \STATE $A_t
  \gets \mathrm{ResolveGraphAssertions}
  (G_t',\widehat{O}_t,\langle e,a,c\rangle$)
  \label{alg:expansion:assertions}

  \FOR{each affected target $u=\langle e,a\rangle$}

    \STATE $W_t(u)
    \gets \mathrm{ComputeWitnessSupport}(u,G_t')$
    \label{alg:expansion:support}

    \IF{$|W_t(u)|\geq2$}
      \STATE $\mathrm{UpdateConsensus}(u,W_t(u),G_t')$
    \ELSE
      \STATE $\mathrm{DeactivateConsensus}(u,G_t')$
    \ENDIF

  \ENDFOR
  \label{alg:expansion:consensus}

  \STATE $V_t
  \gets \mathrm{UpdateValidationRecords}
  (H_t,W_t,P_t,G_t')$
  \label{alg:expansion:validation}

  \STATE $Z_t
  \gets \mathrm{AssignOperationalScopes}
  (A_t,V_t,P_t)$
  \label{alg:expansion:scopes}

  \STATE $T_t
  \gets \mathrm{MaterializeTrustedView}
  (A_t,V_t,P_t)$
  \label{alg:expansion:trusted}

  \STATE $G_t
  \gets \mathrm{PublishVersion}
  ({G_t',A_t,V_t,Z_t},T_t)$
  \label{alg:expansion:publish}

\ENDFOR

\STATE \textbf{return} $G_t,T_t$
\label{alg:expansion:return}

\end{algorithmic}
\end{algorithm}

Algorithm~\ref{alg:expansion} defines the governance procedure by combining claim representation, validation, multi-witness corroboration, and versioned evolution into a single workflow for each ingestion or governance batch.
A batch may introduce new evidence and extraction outputs, add validation grounds, modify the active policy, or revoke previously active sources or decisions.

Line~\ref{alg:expansion:input} specifies the initial audit state \(G_0\) and the ordered sequence
of batches \(B=(b_1,b_2,\ldots)\) processed by the procedure. 
Line~\ref{alg:expansion:output} identifies the outputs: the versioned audit state \(G_t\) and the corresponding trusted view \(T_t\) produced at each iteration.

Line~\ref{alg:expansion:foreach} starts the iteration over the ordered batches.
Line~\ref{alg:expansion:unpack} decomposes the current batch into new evidence $E_t$, extraction records $X_t$, human validation grounds $H_t$, revocations $R_t$, and the active policy $P_t$.

Lines~\ref{alg:expansion:predictions}--\ref{alg:expansion:normalize} implement the claim-representation model of Section~\ref{sec:primitives}: \textsc{PreservePredictions} retains each
run-level output and its provenance, and \textsc{Normalize} resolves evidence references and ATT\&CK identifiers.

Line~\ref{alg:expansion:state} applies activations, supersessions, and revocations, following the versioning model of Section~\ref{sec:expansion}, so that assertion resolution and support recomputation operate on the active state.

Line~\ref{alg:expansion:assertions} then associates the normalized Predictions with the configuration-level key $\langle e,a,c\rangle$, yielding $A_t$, the set of GraphAssertions active at version~$t$.

Lines~\ref{alg:expansion:support}--\ref{alg:expansion:consensus} implement the multi-witness corroboration mechanism of Section~\ref{sec:multiwitness}. For each affected target, \textsc{ComputeWitnessSupport} derives the active setup-deduplicated support set $W_t(u)$. The corresponding ConsensusAssertion is updated when at least two distinct setups support the target and deactivated otherwise.

Lines~\ref{alg:expansion:validation}--\ref{alg:expansion:trusted} apply the validation and trust model of Section~\ref{sec:normalisation}. \textsc{UpdateValidationRecords} combines gold, analyst, and policy-compliant multi-witness grounds; \textsc{AssignOperationalScopes} assigns each GraphAssertion to one exclusive scope; and \textsc{MaterializeTrustedView} derives the active
trusted view.

Finally, Line~\ref{alg:expansion:publish} publishes the new GraphVersion without deleting prior observations or governance decisions.
\section{The \tool{} Framework}\label{sec:framework}

Section~\ref{sec:methodology} defined claim governance independently of any storage technology. This section describes its realization in \tool{}. We map the methodological objects to a claim-centric graph schema, present the processing architecture, and explain how the audit graph, trusted view, evidence granularity, versioning, and revocation are implemented.

\subsection{From Methodology to Graph Representation}\label{sec:design}
The evaluated graph directly implements the three aggregation levels of Section~\ref{sec:primitives}. Persistent observations and derived summaries become first-class entities; validation is represented through explicit, queryable metadata and relations attached to the governed assertions; and the trusted view is derived from their active state. Table~\ref{tab:method_mapping} summarizes the mapping.

\begin{table*}[t]
\caption{Mapping from the methodology to the evaluated \tool{} representation.}
\label{tab:method_mapping}
\centering
\footnotesize
\renewcommand{\arraystretch}{1.08}
\setlength{\tabcolsep}{4pt}
\begin{tabularx}{\textwidth}{@{}p{2.6cm}p{3.3cm}
>{\raggedright\arraybackslash}X
>{\raggedright\arraybackslash}X@{}}
\toprule
\textbf{Methodological object} & \textbf{Graph realization} & \textbf{Purpose} & \textbf{Lifecycle property} \\
\midrule
Evidence unit & Sentence or EvidenceSpan & Preserve the native textual support of a mapping & Never refined beyond source granularity \\
Run-level observation & Prediction node & Record one immutable model output and its full run provenance & Untrusted by default \\
Assertion configuration & ExtractionSetup plus prompting method & Define the aggregation key shared across repeated runs and seeds & RAG and RAG+FSP remain distinct assertion configurations without becoming separate setup witnesses \\
Configuration-level claim & GraphAssertion node & Represent one normalized $\langle e,a,c\rangle$ assertion & Reused across repeated runs of the same configuration \\
Cross-setup sum\-ma\-ry & ConsensusAssertion node & Materialize $\langle e,a\rangle$ targets supported by at least two distinct setups & Measures setup-deduplicated corroboration, not automatic trust \\
Validation event & Trust-event metadata and relations & Record gold, analyst, or multi-witness validation grounds and version & Basis for scope assignment and trusted-view inclusion \\
Versioned state & ImportBatch and GraphVersion & Identify when observations and governance decisions became active & Reconstructible audit history \\
Trusted view & Materialized GraphAssertion set & Expose only active gold or validated assertions & May expand or contract across versions \\
\bottomrule
\end{tabularx}
\end{table*}

Predictions remain run-specific. Repeated seeds with the same assertion configuration may map to a single configuration-level GraphAssertion, whereas RAG and RAG+FSP produce distinct assertion configurations under the same parent ExtractionSetup. GraphAssertions associated with different setups may then converge on a single ConsensusAssertion. Trust remains attached to GraphAssertions through recorded validation grounds, and a ConsensusAssertion provides multi-witness evidence without automatically becoming a trusted fact.

In the current implementation, a validation event is logically explicit and queryable, but is not required to be a separate node label. Its type, supporting object, policy-relevant evidence, graph version, and active state are represented through trust metadata and relations associated with the governed assertion. This preserves the evaluated artifact while enforcing the methodological invariant of Section~\ref{sec:trust}.

\subsection{High-Level Architecture}\label{sec:architecture}

The framework implements the lifecycle through five interconnected macro-layers, as shown in Table~\ref{tab:kg_layers} and Figure~\ref{fig:bpmn}. 
Figure~\ref{fig:bpmn} complements Figure~\ref{fig:method_blocks} by showing how the six governance stages are realized across the architectural layers of \tool{}. Table~\ref{tab:kg_layers} summarizes the core entities, responsibilities, and methodological roles of those layers.

The layers are arranged according to the dependencies of the governance process: evidence is preserved; extraction provenance is captured; outputs are normalized and grounded; configuration-level assertions and cross-setup consensus summaries are constructed; and versioned views are published.

\begin{table*}[t]
\caption{\tool{} layers and the methodological responsibilities they realize.}
\label{tab:kg_layers}
\centering
\footnotesize
\renewcommand{\arraystretch}{1.08}
\setlength{\tabcolsep}{4pt}
\begin{tabularx}{\textwidth}{@{}p{2.7cm}p{4.1cm}
>{\raggedright\arraybackslash}X
>{\raggedright\arraybackslash}X@{}}
\toprule
\textbf{Layer} & \textbf{Core entities} & \textbf{Responsibility} & \textbf{Methodological concept} \\
\midrule
Data and Evidence & Dataset, SourceFile, Report, Sentence, EvidenceSpan & Preserve reports and their native evidence units & Evidence preservation and granularity \\
Extraction Pro\-ve\-nan\-ce & ExtractionSetup, LLMRun, PromptTemplate, RetrievedContext, Prediction & Record how every run-level output was produced & Immutable Prediction and producer traceability \\
Normalization and Grounding & AttackTactic, AttackTechnique, MalontClass & Resolve comparable evidence--technique targets and ontology links & Normalized aggregation key \\
Claims and Va\-li\-da\-tion & GraphAssertion, ConsensusAssertion, gold instances, trust metadata & Represent configuration-level claims, materialize setup-deduplicated corroboration, and record validation grounds & Trust policy and validation provenance \\
Evolution and Governance & ImportBatch, GraphVersion, revocation records, active-state markers & Reconstruct versions and derive current views & Auditable evolution and non-destructive revocation \\
\bottomrule
\end{tabularx}
\end{table*}

\begin{figure*}[t]
\centering
\resizebox{\textwidth}{!}{%
\begin{tikzpicture}[
  font=\scriptsize,
  event/.style={circle,draw,thick,minimum size=10mm,inner sep=1pt,align=center,fill=black!5},
  endevent/.style={circle,draw,very thick,minimum size=10mm,inner sep=1pt,align=center,fill=green!8},
  task/.style={rounded corners=3pt,draw,thick,minimum height=10mm,align=center,fill=blue!5},
  data/.style={draw,densely dashed,align=center,fill=black!3,minimum height=8mm},
  flow/.style={-{Latex[length=2.2mm]},thick},
  assoc/.style={-{Latex[length=1.6mm]},densely dashed,draw=black!55},
  lanelbl/.style={font=\scriptsize\bfseries,text=black!70,align=center}
]
\node[event] (start) at (0.5,0) {CTI\\Report};
\node[task,text width=18mm] (seg) at (2.8,0) {Preserve native\\evidence units};
\node[task,text width=22mm] (extract) at (6.0,0) {Run assertion\\configuration};
\node[data,text width=18mm] (ctx) at (6.0,2.0) {RetrievedContext};
\node[data,text width=18mm] (pred) at (6.0,-2.0) {Prediction};
\node[task,text width=22mm] (norm) at (9.5,0) {Normalize evidence\\and ATT\&CK ID};
\node[task,text width=21mm] (claim) at (12.7,0) {Resolve\\GraphAssertion};
\node[task,text width=22mm] (valid) at (15.9,0) {Update support and\\validation records};
\node[task,text width=21mm] (gov) at (19.2,0.8) {Publish\\GraphVersion};
\node[endevent,text width=13mm] (audit) at (19.2,-1.5) {Audit\\graph};
\node[endevent,text width=13mm] (trusted) at (21.4,0.8) {Trusted\\view};
\draw[flow] (start) -- (seg);
\draw[flow] (seg) -- (extract);
\draw[flow] (extract) -- (norm);
\draw[flow] (norm) -- (claim);
\draw[flow] (claim) -- (valid);
\draw[flow] (valid) -- (gov);
\draw[flow] (gov) -- (trusted);
\draw[flow] (gov) -- (audit);
\draw[assoc] (extract) -- (ctx);
\draw[assoc] (extract) -- (pred);
\draw[assoc] (pred.east) -- (norm.south);
\begin{scope}[on background layer]
  \fill[black!4,rounded corners=2pt]  (-0.4,-3.0) rectangle (4.0,3.1);
  \fill[blue!3,rounded corners=2pt]   (4.1,-3.0) rectangle (7.8,3.1);
  \fill[teal!4,rounded corners=2pt]   (7.9,-3.0) rectangle (11.2,3.1);
  \fill[orange!4,rounded corners=2pt] (11.3,-3.0) rectangle (17.6,3.1);
  \fill[violet!4,rounded corners=2pt] (17.7,-3.0) rectangle (22.4,3.1);
  \foreach \x/\lbl in {1.8/{Data \&\\Evidence},5.95/{Extraction\\Provenance},9.55/{Normalization \&\\Grounding},14.45/{Claims \&\\Validation},20.05/{Evolution \&\\Governance}}
     \node[lanelbl] at (\x,2.7) {\lbl};
  \draw[black!25] (4.05,-3.0) -- (4.05,3.1);
  \draw[black!25] (7.85,-3.0) -- (7.85,3.1);
  \draw[black!25] (11.25,-3.0) -- (11.25,3.1);
  \draw[black!25] (17.65,-3.0) -- (17.65,3.1);
\end{scope}
\end{tikzpicture}}
\caption{Process view of \tool{}. Grounding precedes cross-setup comparison; validation records determine the trusted view, while all observations remain in the audit graph.}
\label{fig:bpmn}
\end{figure*}
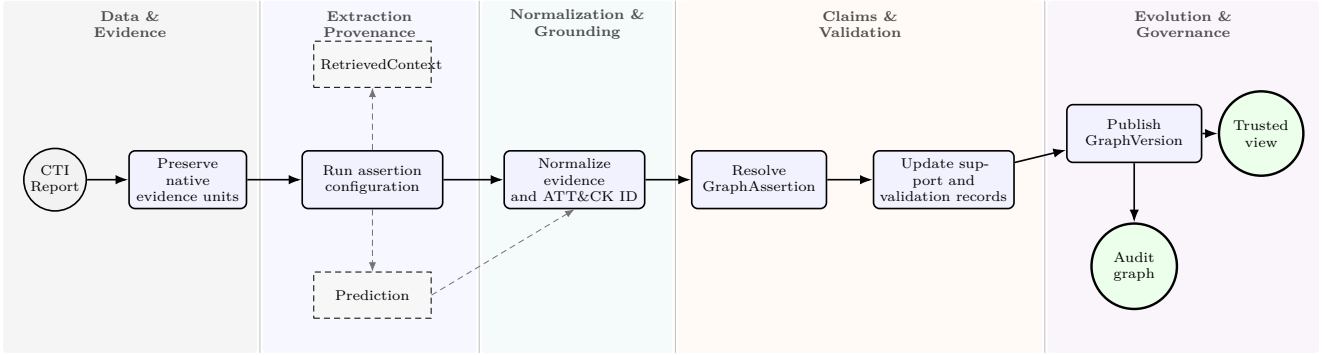

In the \emph{Data and Evidence} layer, reports are segmented while preserving whether the available support is sentence-level or span-level. In \emph{Extraction Provenance}, a replaceable ExtractionSetup is invoked through an assertion configuration, emits Predictions, and records retrieved context separately from source evidence. This distinction allows later queries to inspect what the model considered and what it actually asserted.

The \emph{Normalization and Grounding} layer resolves technique identifiers and canonical evidence units before outputs are compared. ATT\&CK supplies the target tactics and techniques, while MALOnt may provide supporting cybersecurity concepts~\cite{rastogi2020malont}. The \emph{Claims and Validation} layer maps Predictions to configuration-level GraphAssertions, materializes cross-setup support as ConsensusAssertions after deduplication by parent ExtractionSetup, and records the trust metadata that justifies scope assignment. Finally, \emph{Evolution and Governance} assigns ImportBatches and GraphVersions, records revocations, and derives the audit and trusted views.

\subsection{Claim-Centric Knowledge Graph Schema}\label{sec:schema}

The schema is organized around three connected paths. The run-level provenance path reconstructs a model observation:
\[
\begin{aligned}
\text{Report}&\rightarrow\text{EvidenceUnit}\rightarrow\text{Prediction}\\
&\rightarrow\text{LLMRun}\rightarrow\text{ExtractionSetup}.
\end{aligned}
\]
The prompting method associated with the LLMRun completes the logical assertion configuration $c=\langle s,m\rangle$; AssertionConfiguration is therefore a methodological aggregation key, not a separate node class in the evaluated schema.

The configuration-level assertion path links repeated observations to the claim made by one assertion configuration:
\[
\text{Prediction}\rightarrow\text{GraphAssertion}
\rightarrow\text{AttackTechnique}.
\]
The cross-setup governance path materializes corroboration and its trust consequences:
\[
\begin{aligned}
\text{GraphAssertion}&\rightarrow\text{ConsensusAssertion}\\
&\rightarrow\substack{\text{support and}\\\text{trust metadata}}
\rightarrow\text{GraphVersion}.
\end{aligned}
\]

Together, these paths answer different audit questions. The first identifies the text, retrieved context, run, seed, prompting method, and parent setup behind an output. The second shows which normalized assertion that configuration contributes. The third identifies the distinct setups supporting the same target, the agreement level reached, the validation basis used by the policy, and the version in which the resulting scope was active. The complete entity inventory remains in \ref{app:schema} (Table~\ref{tab:core_entities}).

\begin{figure*}[t]
\centering
\includegraphics[width=\textwidth]{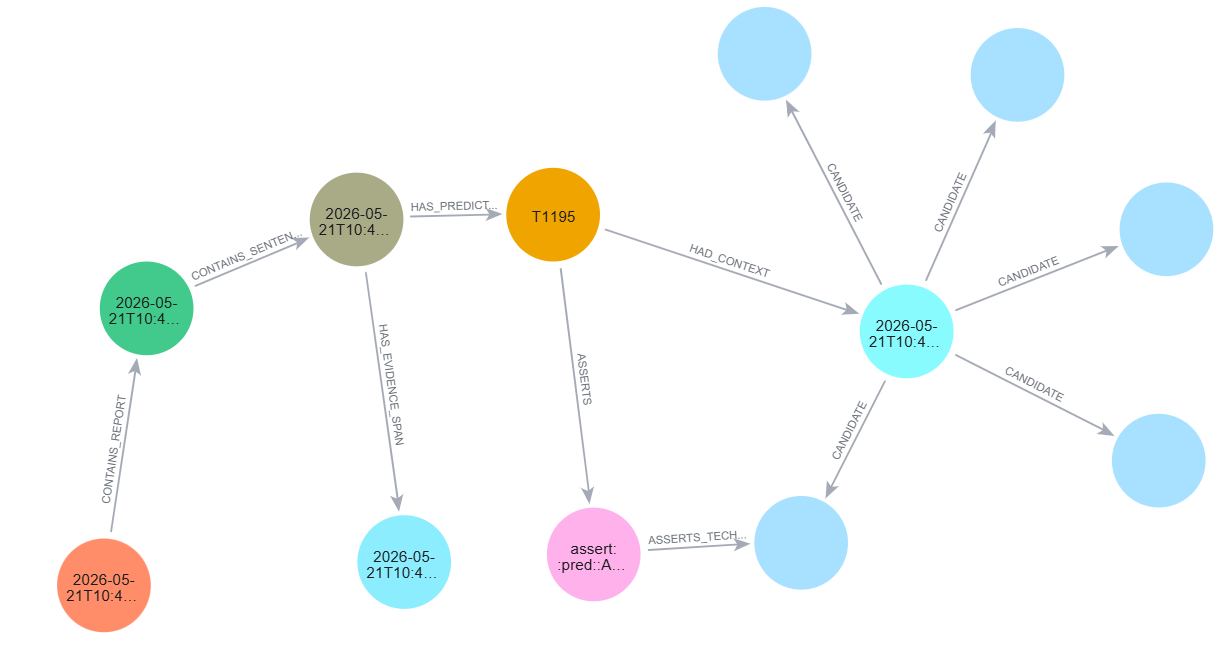}
\caption{Provenance and governance paths from run-level Predictions to configuration-level GraphAssertions and cross-setup ConsensusAssertions. The latter summarize setup-deduplicated corroboration and do not constitute trusted facts on their own.}
\label{fig:kg_audit_path}
\end{figure*}

Each Prediction has one outgoing assertion relation and zero or more relations to RetrievedContext items. Predictions produced by repeated runs or seeds of the same assertion configuration can map to the same GraphAssertion. RAG and RAG+FSP are retained as distinct assertion configurations because the latter augments the common sentence-and-retrieved-context input with five few-shot examples. Both configurations inherit the same parent ExtractionSetup for witness counting. The retrieved items record material made available to the model; they are not treated as source evidence unless the validation record explicitly identifies them as supporting grounds.

GraphAssertions sharing the same normalized evidence--technique target are connected through one ConsensusAssertion when their parent-setup set contains at least two distinct active ExtractionSetups. The ConsensusAssertion stores or exposes the setup-deduplicated support count and family composition needed for agreement queries. Consequently, multiple methods, repeated runs, or repeated seeds under the same parent setup cannot inflate corroboration.

The topology remains stable when new extractors are added. A new setup creates new assertion configurations, runs, Predictions, and GraphAssertions and may create or version ConsensusAssertions, while existing reports, evidence units, ATT\&CK concepts, and prior audit paths remain unchanged. This realizes extractor-independent evolution without conflating ingestion, corroboration, and trust assignment.

\subsection{Evidence Granularity and Canonicalization}\label{sec:granularity}

CTI corpora and extractors may identify evidence at different levels of textual granularity. \tool{} preserves this distinction explicitly. A sentence-level mapping is represented as supported by that sentence; a span-level mapping is represented by the annotated substring and linked to its enclosing sentence. The framework permits upward projection from a span to its sentence for broader retrieval, but never creates a finer trusted span from sentence-level evidence.

Canonicalization proceeds at two related levels. First, Predictions are aggregated into the same GraphAssertion only when their normalized evidence unit, ATT\&CK identifier, parent ExtractionSetup, and prompting method match; run and seed are deliberately excluded from this aggregation key. Second, configuration-level GraphAssertions support the same ConsensusAssertion target only when their normalized evidence unit and ATT\&CK identifier match. At this second level, support is deduplicated by parent ExtractionSetup, so RAG and RAG+FSP variants of the same setup do not create two witnesses.

If a coarser comparison is required, it must be exposed as a separate query view rather than silently rewriting the underlying evidence. This policy preserves the evidential precision of AnnoCTR-like span annotations while remaining compatible with TRAM-like sentence annotations. It also makes the assumptions behind each agreement measurement explicit and queryable.

\subsection{Validation and Trusted-View Materialization}\label{sec:trusted-view}

The audit graph contains all Predictions, GraphAssertions, ConsensusAssertions, gold instances, support relations, trust metadata, and revocation records. The trusted view is the subset of GraphAssertions satisfying the invariant in Section~\ref{sec:trust}. It is derived from recorded validation grounds rather than from an unexplained mutable flag.

Gold validation links a GraphAssertion to the corresponding human annotation. Analyst validation records the reviewer's decision and, when available, the version. For multi-witness validation, the framework uses the associated ConsensusAssertion to recover distinct setup support and the represented generator and retriever families. If this evidence satisfies the active policy, the affected GraphAssertions enter the validated scope. The ConsensusAssertion remains a summary of the inclusive, corroborated target and does not, by itself, become trusted merely because it exists.

Agreement views $C_t^{(2)}$, $C_t^{(3)}$, and $U_t$ are computed from ConsensusAssertion support metadata. They support the precision--recall and unanimity analyses in the evaluation. The experimental four-way decomposition remains unchanged: GraphAssertions are assigned to gold, validated, prediction-only, or deprecated scopes; corroborated and unanimous subsets refine the validated results and can be queried separately.

The downstream view is therefore
\[
\begin{aligned}
T_t=\{g:\;&\operatorname{active}(g,t)\ \land\\
&\operatorname{scope}(g,t)\in\{\mathrm{gold},\mathrm{validated}\}\}.
\end{aligned}
\]
Prediction-only GraphAssertions and all underlying Predictions remain available for audit and review-queue construction but are not silently merged into $T_t$.

\subsection{Versioning and Non-Destructive Revocation}\label{sec:evolution}

Each ingestion or governance batch creates an ImportBatch and publishes a GraphVersion. Nodes and relations record the batch in which they were introduced and, where relevant, the versions in which they are active. Stable identifiers allow the framework to reuse existing reports, evidence units, ATT\&CK concepts, GraphAssertions, and consensus targets without rewriting earlier paths.

Revocation is represented as a new governance record. If a dataset, setup, assertion configuration, run, or validation decision becomes unreliable, the corresponding object is deactivated rather than deleted. The framework then recomputes affected ConsensusAssertion support and reevaluates policy-derived validation records. GraphAssertions that lose their final qualifying record leave the trusted view in the new GraphVersion, while previous versions still show why they were formerly trusted.

Consequently, \tool{} provides two complementary guarantees: the audit history is append-only and reconstructible, while the active state remains responsive to corrections. This distinction supports non-destructive revocation more accurately than a model in which knowledge can only grow.

\subsection{Framework Guarantees and Evaluation Boundary}
\label{sec:framework-guarantees}

The framework guarantees, by construction, that raw model outputs are preserved; configuration-level GraphAssertions remain linked to evidence and producer provenance; trust changes are backed by explicit, queryable validation records; setup support is deduplicated before corroboration; and previous graph states remain reconstructible. These are structural properties of the representation and governance procedure.

Other properties require empirical evaluation. In particular, the framework does not assume that distinct setups are independent or that a fixed support threshold guarantees correctness. The experiments therefore characterize output diversity across same- and cross-generator-family
setup pairs, examine the precision--recall trade-off across support levels, verify that the schema and audit paths remain stable as setups are ingested, and determine which SOC-oriented questions become directly answerable.
Section~\ref{sec:evaluation} evaluates these questions under the concrete
instantiation described in Section~\ref{sec:instantiation}.

\section{Evaluation}
\label{sec:evaluation}

We evaluate whether \tool{} realizes the four properties posed in Section~\ref{sec:intro}: auditable evolution (RQ1), extractor-independent governance and informative witness diversity (RQ2), controlled promotion to trusted knowledge (RQ3), and graph-native operational utility (RQ4). The evaluation separates two kinds of evidence. \emph{Conformance evidence} checks whether the implementation preserves the structural invariants defined in Sections~\ref{sec:methodology}--\ref{sec:framework}; \emph{empirical evidence} measures properties that are not guaranteed by construction, including the behavioral diversity of heterogeneous extraction setups and the precision--recall trade-off induced by multi-witness policies.

The upstream prediction protocol follows the 2025 USENIX Security SoK on automated TTP extraction~\cite{buchel2025sok}.\label{sec:sok} The SoK pipeline supplies the Predictions governed by \tool{} and is not treated as a competing governance system. We implement the graph in Neo4j~5~\cite{neo4j2024}, query it with Cypher, and load Neo4j-safe CSV files, while keeping the report text in content-hashed JSONL. All reported values are derived from the published extraction outputs and graph snapshot; no additional live LLM call is required.

\begin{table*}[t]
\caption{Evaluation questions, evidence, and primary measurements.}
\label{tab:rq_evidence}
\centering
\small
\renewcommand{\arraystretch}{1.08}
\setlength{\tabcolsep}{4pt}
\begin{tabularx}{\textwidth}{@{}p{1.0cm}p{3.1cm}>{\raggedright\arraybackslash}X>{\raggedright\arraybackslash}X@{}}
\toprule
\textbf{RQ} & \textbf{Evaluation focus} & \textbf{Evidence} & \textbf{Primary measurements} \\
\midrule
RQ1 & Auditable evolution & Six cumulative GraphVersions produced by ingesting S1--S6 & schema delta, provenance closure, scope disjointness, version-localisability, graph growth \\
RQ2 & Setup compatibility and witness diversity & Closed $2\times3$ retriever--generator matrix & successful ingestion, mean ATT\&CK reach, reference F1, Jaccard output diversity, correctness of cumulative agreement \\
RQ3 & Controlled promotion & Gold, validated, prediction-only, and deprecated GraphAssertion scopes & validation-record conformance, leakage, agreement views, gold-aligned precision and recall \\
RQ4 & Operational utility & Audit queries Q1--Q7 over the v6.0 snapshot and a flat-output baseline & direct answerability, required enrichment or reprocessing, counterfactual setup dependencies \\
\bottomrule
\end{tabularx}
\end{table*}

\subsection{Experimental Design and Measurement Protocol}\label{sec:instantiation}\label{sec:extraction}

\subsubsection{Corpora and extraction setups}

We exercise the methodology on the two public corpora used by the upstream evaluation: TRAM~v2~\cite{ctid2023tram}, which supplies sentence-level ATT\&CK annotations, and AnnoCTR~\cite{lange2024annoctr}, which supplies span-level annotations. Both are ingested in their entirety, for a total of 65 reports and 5{,}303 sentences. \tool{} preserves the native evidence granularity according to Section~\ref{sec:granularity}.

The ontology layer is built from the Enterprise, Mobile, and ICS STIX bundles of MITRE ATT\&CK v19.1, retrieved on 22 May 2026. This version identifies the ontology snapshot imported into the graph. The retrieval knowledge base and dataset-specific label sets are instead inherited unchanged from the frozen upstream SoK pipeline; they were not rebuilt from the v19.1 bundles. The three STIX bundles are included among the audit path's source files so that the provenance of ontology grounding remains queryable.

The controlled extraction matrix contains two retrievers, E5-large-v2~\cite{wang2022e5} and GTE-Qwen2-7B-Instruct~\cite{alibaba2024gte}, and three generators, Llama-3.1-8B-Instruct~\cite{meta2024llama}, Mistral-7B-Instruct-v0.3~\cite{jiang2023mistral}, and Phi-3.5-mini-instruct~\cite{abdin2024phi3}. Each retriever--generator cell defines one parent ExtractionSetup. The six setups are ingested in the order S1--S6, producing the cumulative graph states v1.0--v6.0 shown in Table~\ref{tab:setup_map}. The order exercises versioned ingestion; it is not interpreted as a randomized treatment order.

Each setup is executed on both datasets, with RAG and RAG+FSP prompting, and with three recorded seeds. Thus, each $(\text{setup},\text{dataset},\text{method},\text{seed})$ tuple defines one LLMRun, yielding
\[
6\times2\times2\times3=72\ \text{LLMRuns}.
\]
Both prompting methods receive the source sentence and the retrieved context. RAG+FSP additionally includes five few-shot examples in the prompt, whereas RAG does not. Native AnnoCTR span annotations are preserved at their source granularity; model outputs are not interpreted as supporting a finer span that the extractor did not identify. RAG and RAG+FSP remain distinct assertion configurations $c=\langle s,m\rangle$, but both inherit the same parent ExtractionSetup and therefore contribute at most one setup witness to $W_t(u)$. The generator uses zero-temperature greedy decoding with a frozen retriever. For each sentence, the retriever first returns the top 100 candidates over the frozen knowledge base; candidates outside the dataset label set are removed, and the first $k_{\mathrm{ret}}=5$ remaining items are supplied as retrieved context. This retrieval depth is fixed across all setups and runs. The three recorded seeds produce identical outputs ($\sigma=0$). They are retained for reproducibility and provenance, not treated as independent statistical replicates.

The retrieval depth $k_{\mathrm{ret}}=5$, the five few-shot examples used by RAG+FSP, and the agreement threshold $k_t(u)$ are distinct quantities.

\begin{table}[t]
\caption{Setup--state mapping for the experimental instantiation.}
\label{tab:setup_map}
\centering
\footnotesize
\setlength{\tabcolsep}{2pt}
\resizebox{\columnwidth}{!}{%
\begin{tabular}{@{}llllll@{}}
\toprule
\textbf{Setup} & \textbf{State} & \textbf{Retriever} & \textbf{Generator} & \textbf{Cell} & \textbf{Role} \\
\midrule
S1 & v1.0 & GTE-Qwen2 & Llama-3.1 & (GTE,Llama) & initial state \\
S2 & v2.0 & E5 & Mistral & (E5,Mistral) & second family \\
S3 & v3.0 & GTE-Qwen2 & Mistral & (GTE,Mistral) & enables $k=3$ \\
S4 & v4.0 & E5 & Llama-3.1 & (E5,Llama) & closes $2\times2$ \\
S5 & v5.0 & E5 & Phi-3.5 & (E5,Phi-3.5) & third family \\
S6 & v6.0 & GTE-Qwen2 & Phi-3.5 & (GTE,Phi-3.5) & closes $2\times3$ \\
\bottomrule
\end{tabular}}
\end{table}

\subsubsection{Units of analysis}

The evaluation keeps the three methodological levels separate.

\begin{itemize}
\item A \emph{Prediction} is one run-level output and is used to verify provenance and reproducibility.
\item A \emph{GraphAssertion} $g=\langle e,a,c\rangle$ is a configuration-level governed assertion and is the unit used for operational scopes and trusted-view exposure.
\item A \emph{canonical target} $u=\langle e,a\rangle$ is configuration-independent. Support $k_t(u)=|W_t(u)|$, ConsensusAssertions, and agreement levels are computed at this target level after deduplication by parent ExtractionSetup.
\end{itemize}

This distinction prevents the count of ConsensusAssertions from being interpreted as a fraction of configuration-level GraphAssertions: the two values describe different aggregation levels. For analysis, let
\[
C_t^{(k)}=\{u:k_t(u)\geq k\}
\]
be the target-level agreement view and let
\[
E_t^{(k)}=\{g\in A_t:\operatorname{target}(g)\in C_t^{(k)}\}
\]
be its GraphAssertion exposure view. Target-level support and GraphAssertion-level trusted-view size are reported separately.

\paragraph{Operational policy and analytical agreement views}
We distinguish the operational trust policy used to materialize the reported trusted view from stricter agreement-conditioned views used only for analysis. In the evaluated snapshot, a GraphAssertion enters the gold scope when its document-level target is backed by a trusted corpus annotation of the same document. A non-gold GraphAssertion enters the validated scope when its target is supported by at least two distinct active parent ExtractionSetups, after deduplication of RAG/RAG+FSP configurations and repeated seeds. This operational rule is applied uniformly across v1.0--v6.0. The stricter $k\geq3$ and unanimity views evaluated below are counterfactual analytical views derived from the same ConsensusAssertion support metadata; they do not redefine the operational scope assignment used to obtain the trusted-view counts in Table~\ref{tab:headline}.

\subsubsection{Correctness and diversity measurements}

Gold-aligned precision and recall follow the document-level matching policy of the upstream evaluation. Gold annotations from both corpora are normalized to document-level tuples $\langle d,a\rangle$, where $d$ is the report and $a$ the normalized ATT\&CK identifier; the 824 gold instances yield 663 distinct document-level tuples. Each GraphAssertion $g$ in an exposure view $V$ is likewise projected to its document-level tuple $\langle d(g),a(g)\rangle$, and duplicate tuples are counted once. Let $D(V)$ denote the resulting tuple set, $D^{*}(V)\subseteq D(V)$ the subset whose documents carry at least one gold annotation, and $Y$ the set of gold document-level tuples. Per corpus,
\[
P(V)=\frac{|D^{*}(V)\cap Y|}{|D^{*}(V)|},
\qquad
R(V)=\frac{|D(V)\cap Y|}{|Y|},
\]
and the reported aggregates are macro-averages across TRAM~v2 and AnnoCTR; per-corpus results are given where available. Two of the 65 reports (one per corpus) contain no gold annotation; assertions for these documents remain in the graph and in the reported view sizes, but are excluded from the precision denominator. View sizes remain reported in GraphAssertions, whereas correctness is computed over deduplicated document-level tuples; the two units must not be mixed. Document-level alignment is deliberately coarser than sentence- or span-level recovery: an exposed assertion counts as gold-aligned when its technique is annotated anywhere in the same document, so the reported precision and recall are upper bounds with respect to finer-grained matching. The native sentence-level (TRAM~v2) and span-level (AnnoCTR) annotations remain available in the source corpora and are never projected to finer artificial units (Section~\ref{sec:granularity}). Gold membership is also used as a conformance ground for the gold operational scope, which is likewise assigned at document level; the latter is not interpreted as an extractor-performance result.

Output diversity between two setup cells is measured with Jaccard distance over their predicted ATT\&CK-ID sets:
\[
d_J(A,B)=1-\frac{|A\cap B|}{|A\cup B|}.
\]
A larger value indicates less overlap. Jaccard distance measures behavioral diversity, not statistical independence of errors. Accordingly, the evaluation uses \emph{distinct}, \emph{heterogeneous}, and \emph{diverse} witnesses rather than assuming independent draws.

\subsection{Artifact Overview}\label{sec:headline}

Table~\ref{tab:headline} reports the v6.0 graph inventory; counts on different rows may refer to different aggregation levels. The counts illustrate the three aggregation levels: 82{,}260 run-level Predictions are consolidated into 27{,}420 configuration-level GraphAssertions, while 5{,}410 ConsensusAssertions materialize configuration-independent targets supported by at least two distinct parent setups. The implementation contains one evidence record for each of the 5{,}303 extractor-input sentences; this technical count is not used as a count of native AnnoCTR spans and does not imply that sentence-level TRAM annotations were refined to artificial spans. Policy and matching retain the native corpus granularity described above.

The 68 source files comprise the 65 CTI report files (34 from AnnoCTR and 31 from TRAM~v2) and the three MITRE ATT\&CK STIX bundles used as ontology sources (Enterprise, Mobile, and ICS). The latter belong to the audit path because they provide the provenance of the grounding layer.\footnote{In the published snapshot, corpus membership and file-level provenance are recorded as properties of Report, Sentence, and LLMRun nodes and of the ontology import rather than as separate Dataset or SourceFile nodes; the count enumerates the 65 report files and the three STIX bundles.} MALOnt enrichment is not included in this count because it remains outside that path.

\begin{table*}[t]
\caption{Artifact inventory for graph v6.0.}
\label{tab:headline}
\centering
\small
\begin{tabular}{@{}lr@{}}
\toprule
\textbf{Entity} & \textbf{Count} \\
\midrule
Dataset / Report / Sentence & 2 / 65 / 5{,}303 \\
Implementation evidence records & 5{,}303 \\
Source files / ImportBatch & 68 / 6 \\
LLMRun / ExtractionSetup / GraphVersion & 72 / 6 / 6 \\
Prediction & \textbf{82{,}260} \\
RetrievedContext / CONSIDERED edges & 89{,}706 / 140{,}630 \\
AttackTactic & 21 \\
AttackTechnique (1,149 from v19.1 STIX + 101 observed-ID placeholders) & 1,250 \\
ATTACK\_RELATIONSHIP edges & 21{,}324 \\
MalontClass / MALONT\_RELATION & 75 / 60 \\
GraphAssertion (configuration level) & \textbf{27{,}420} \\
ConsensusAssertion (target level, $k\geq2$) & \textbf{5{,}410} \\
Gold annotation instances & \textbf{824} \\
SUPPORTS / AGREES\_WITH / DISAGREES\_WITH edges & 15{,}120 / 16{,}991 / 68{,}231 \\
Trusted-view GraphAssertions (gold $\cup$ validated) & \textbf{20{,}209} \\
\bottomrule
\end{tabular}
\end{table*}

Of the 1,250 \texttt{AttackTechnique} nodes, 1,149 are imported from the three ATT\&CK v19.1 STIX bundles. The remaining 101 are placeholder nodes for identifiers observed in model Predictions or gold annotations but absent from the current ontology snapshot. Such identifiers may correspond to unsupported model outputs, techniques deprecated or removed between ATT\&CK releases, or annotations based on an earlier catalog. Placeholder nodes retain the observed \texttt{attack\_id} and are marked \texttt{observed\_not\_in\_current\_stix\_bundle}. They ensure that every assertion retains a queryable target rather than silently discarding outputs that cannot be grounded in the current bundles.

\subsection{RQ1: Schema-Stable and Auditable Evolution}\label{sec:coverage}\label{sec:evolution-empirical}\label{sec:schema-util}\label{sec:attack-coverage}

RQ1 asks whether successive setups can be incorporated without changing the schema or breaking existing provenance paths. The schema and ingestion procedure establish the required path shape by construction; the six-version ladder verifies conformance for all ingested data.

At every ImportBatch, we check: (i) path completeness from each GraphAssertion to evidence, run, setup, and report; (ii) uniqueness of the outgoing ASSERTS relation from each Prediction; (iii) disjointness of the four operational scopes; (iv) version-localisability through ImportBatch and GraphVersion; and (v) evidence-granularity safety. Across v1.0--v6.0, evidence coverage and path completeness are 100\%, no dangling provenance reference or duplicate assertion relation is observed, and the operational scopes remain disjoint. All entity types materialized in the snapshot are exercised at v6.0; Dataset, SourceFile, PromptTemplate, and EvidenceSpan are realized as node properties or as external source artifacts in this instantiation (\ref{app:schema}). Ingesting S1--S6 introduces no node labels, relation types, or schema constraints beyond those already present in v1.0.

\begin{table*}[t]
\caption{Evolution and conformance across cumulative graph states. A dash indicates that corroboration is not defined by a single active setup.}
\label{tab:evolution_conformance}
\centering
\small
\setlength{\tabcolsep}{5pt}
\resizebox{\textwidth}{!}{%
\begin{tabular}{@{}lrrrrccc@{}}
\toprule
\textbf{State} & \textbf{Active setups} & \textbf{GraphAssertions} & \textbf{ATT\&CK IDs} & \textbf{$C_t^{(2)}$ targets} & \textbf{Path closure} & \textbf{Scope overlap} & \textbf{Schema delta} \\
\midrule
v1.0 & 1 & 5{,}249 & 120 & -- & 100\% & 0 & 0 \\
v2.0 & 2 & 11{,}930 & 133 & 1{,}539 & 100\% & 0 & 0 \\
v3.0 & 3 & 19{,}065 & 137 & 3{,}952 & 100\% & 0 & 0 \\
v4.0 & 4 & 23{,}961 & 137 & 4{,}810 & 100\% & 0 & 0 \\
v5.0 & 5 & 25{,}625 & 138 & 5{,}088 & 100\% & 0 & 0 \\
v6.0 & 6 & 27{,}420 & 139 & 5{,}410 & 100\% & 0 & 0 \\
\bottomrule
\end{tabular}}
\end{table*}

Figure~\ref{fig:kg_evolution} complements the conformance table with descriptive growth signals: panel~(a) tracks configuration-level GraphAssertions, panel~(b) distinct ATT\&CK IDs, panel~(c) the inclusive target-level corroborated view $C_t^{(2)}$, and panel~(d) the exclusive non-gold strict-unanimity component. GraphAssertion volume increases from 5{,}249 to 27{,}420, while observed ATT\&CK reach increases from 120 to 139 IDs. The target-level corroborated view $C_t^{(2)}$ becomes available at v2.0 and grows to 5{,}410 targets at v6.0. Panel~(d) starts at v3.0 because, with at most two active setups, unanimity does not define a stricter component than the $k\geq2$ corroboration floor. These target counts are not divided by the configuration-level GraphAssertion counts because they describe different units.

\begin{figure*}[!b]
\centering
\includegraphics[width=\textwidth]{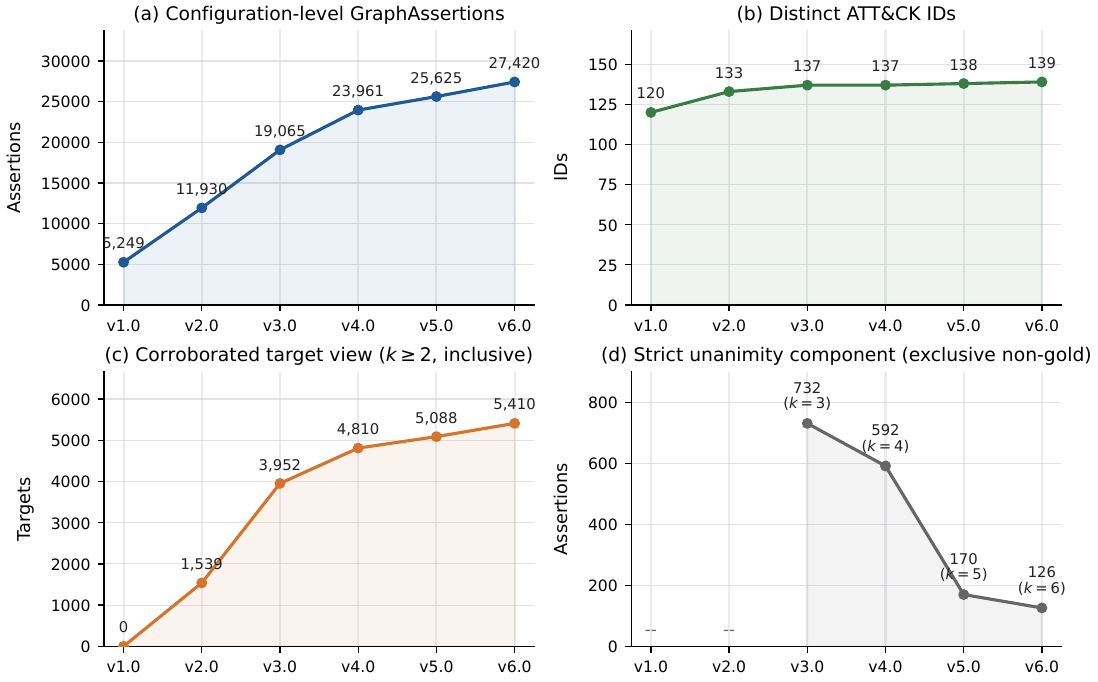}
\caption{Evolution across v1.0--v6.0 of assertions, ATT\&CK reach, corroborated targets, and strict unanimity.}
\label{fig:kg_evolution}
\end{figure*}

\smallskip\noindent\textbf{Preservation rather than monotonic trust.}\label{sec:panel-a-volume}
The ingestion schedule contains additions but no revocation batch; consequently, its active trusted view grows from 2{,}850 to 20{,}209 GraphAssertions. This observation does not imply that the methodology is monotonic: Sections~\ref{sec:expansion} and~\ref{sec:evolution} allow a later governance batch to contract the active view while retaining earlier records. RQ1 establishes that this additive schedule preserves existing audit paths and requires no schema migration.

\smallskip\noindent\textbf{Reach and corroboration.}\label{sec:panel-b-reach}\label{sec:panel-c-consensus}\label{sec:panel-d-strong}
ATT\&CK reach follows $120\to133\to137\to137\to138\to139$, whereas the inclusive corroborated target view follows $0\to1{,}539\to3{,}952\to4{,}810\to5{,}088\to5{,}410$. The near-plateau in reach and continued growth in corroboration indicate that later setups mainly add alternative support for already observed targets. This is the regime required for the multi-witness analyses of RQ2 and RQ3.

\paragraph{RQ1 finding}
All six heterogeneous setups are ingested with the same schema, loader, and provenance path. Every checked GraphAssertion remains evidence- and producer-traceable, previous states remain version-localizable, and no schema element is added between v1.0 and v6.0. The evaluation therefore supports auditable, schema-stable evolution for the observed additive schedule.


\subsection{RQ2: Extractor-Independent Governance and Witness Diversity}\label{sec:multi-llm-value-empirical}

RQ2 has two parts. First, extractor-independent governance requires the same representation and policy to accept heterogeneous setups. Second, witness diversity asks whether heterogeneous generators provide an informative validation signal beyond repeated or closely related configurations.

\subsubsection{Heterogeneous-setup compatibility}

The six retriever--generator cells are imported through the same node labels, relations, canonicalization rules, and trust policy. RAG and RAG+FSP remain distinguishable as assertion configurations, and the three deterministic seed records remain traceable, yet neither methods nor seeds inflate $W_t(u)$. The deferred (GTE-Qwen2, Phi-3.5) cell is admitted at v6.0 without schema or loader modification. These observations support the structural, extractor-independent part of RQ2.

\subsubsection{Upstream prediction profile}\label{sec:factorial}\label{sec:bias}

Table~\ref{tab:factorial_2x3} characterizes the prediction substrate governed by \tool{}. Mean ATT\&CK reach and doc-level F1 are descriptive upstream measurements rather than governance outcomes. The non-integer reach values are aggregated means over the evaluation strata used by the reused protocol, not graph-wide counts of distinct IDs. The table pools RAG and RAG+FSP results and is therefore not interpreted as a causal estimate of the effect of few-shot prompting. Both methods receive the source sentence and the retrieved context; RAG+FSP additionally includes five few-shot examples in the prompt.

\begin{table*}[!hb]
\centering
\small
\caption{Mean ATT\&CK reach and reference doc-level F1 for the closed $2\times3$ setup matrix.}
\label{tab:factorial_2x3}
\label{tab:factorial_2x3_f1}
\setlength{\tabcolsep}{5pt}
\begin{tabular}{@{}llrrr@{}}
\toprule
 & & \multicolumn{3}{c}{\textbf{Generator family}} \\
\cmidrule(lr){3-5}
\textbf{Retriever} & & \textbf{Llama} & \textbf{Mistral} & \textbf{Phi-3.5} \\
\midrule
\multirow{2}{*}{E5} & mean ATT\&CK reach & 60.75 & 71.00 & 58.50 \\
 & F1 & 0.44 & 0.47 & 0.39 \\
\addlinespace
\multirow{2}{*}{GTE-Qwen2} & mean ATT\&CK reach & 65.75 & 75.50 & 61.25 \\
 & F1 & 0.47 & 0.47 & 0.38 \\
\bottomrule
\end{tabular}
\end{table*}

Aggregates reported in the text are computed from the unrounded stratum-level values; the cells in Table~\ref{tab:factorial_2x3} are rounded to two decimal places.

Mistral has the broadest mean reach in both retriever rows, followed by Llama and Phi-3.5. The same ordering is not mirrored exactly by doc-level F1: the generator-family means are 0.455 for Llama, 0.471 for Mistral, and 0.384 for Phi-3.5, while the retriever-row F1 means differ by 0.005. We therefore treat breadth and reference F1 as complementary descriptive properties. The design does not isolate whether the narrower Phi-3.5 output originates from scale, training data, instruction tuning, or their interaction.

\subsubsection{Within- and cross-family output diversity}\label{sec:within-cross-family}

The closed $2\times3$ matrix yields 15 unordered setup pairs, each evaluated over four dataset--method strata, for 60 stratified comparisons. The three setup pairs that share a generator family contribute 12 comparisons; the remaining 12 cross-generator-family setup pairs contribute 48 comparisons. The mean Jaccard distance is 0.3184 for pairs sharing the same generator family and 0.4326 for cross-generator-family pairs, a difference of 0.114. Thus, cross-family pairs exhibit \emph{greater output diversity}, not greater raw agreement. The corresponding retriever-axis contrast is reported as $-0.020$, indicating no analogous positive diversity gap. Figure~\ref{fig:within_vs_cross_plot} reports the corresponding group means of the sentence-level Jaccard distance, pooled across the four dataset--method strata ($N=12/48$ for same-/different-generator-family pairs and $N=24/36$ for same-/different-retriever pairs); error bars show $\pm1$ sample standard deviation over the pairwise comparisons and are not confidence intervals for the group means.

\begin{figure}[!hb]
\centering
\includegraphics[width=\columnwidth]{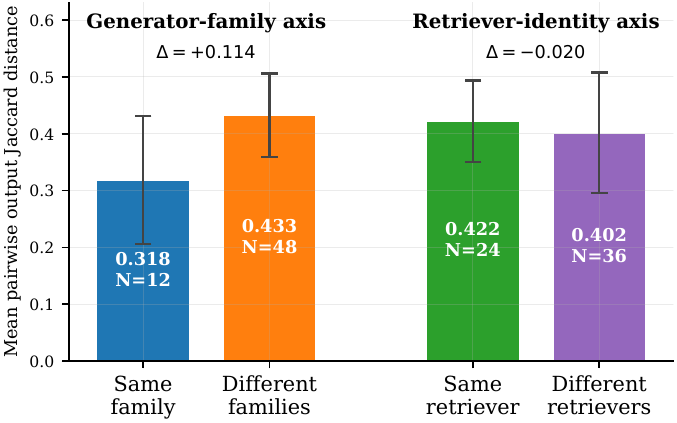}
\caption{Setup-pair output diversity by configuration axis in the closed $2\times3$ matrix. Means are pooled across the four dataset--method strata; error bars show $\pm1$ sample standard deviation across setup pairs.}
\label{fig:within_vs_cross_plot}
\end{figure}

Complementary correctness evidence is supplied by the cumulative unanimity schedule reported in Section~\ref{sec:consensus-precision}. This schedule shows that agreement over an increasingly heterogeneous active set is associated with higher gold-aligned precision: Macro-averaged precision across the two corpora rises from 38.0\% at v1.0 to 90.6\% at v6.0 as the active set expands from one setup and one generator family to all six setups and three generator families. It does not directly compare same-family and cross-family support at fixed $k$, because support threshold, active-set composition, and ingestion order vary jointly. The two analyses therefore provide converging but not controlled evidence: the pairwise comparison establishes behavioral diversity, while the cumulative schedule establishes the correctness associated with agreement across the expanding heterogeneous witness set.

\paragraph{RQ2 finding}
The same governance model accommodates all six setup cells without schema or loader changes. Cross-generator-family pairs exhibit greater behavioral diversity than pairs sharing a generator family, while full-matrix agreement is associated with substantially higher gold-aligned precision along the specified cumulative schedule. These results establish extractor-independent governance and provide supportive evidence that heterogeneous corroboration is informative in this instantiation. They do
not directly estimate a cross-family precision advantage over within-family corroboration at fixed support size and therefore do not support a causal family-diversity claim.


\subsection{RQ3: Controlled Promotion to the Trusted View}\label{sec:trust-policy}\label{sec:consensus-precision}

RQ3 concerns whether every prediction-derived GraphAssertion exposed as trusted has an explicit qualifying validation record. We therefore evaluate policy conformance separately from the empirical correctness of analytical agreement views.

\subsubsection{Operational scopes and policy conformance}

The operational scopes are the four exclusive categories defined in Section~\ref{sec:trust}: gold, validated, prediction-only, and deprecated. Corroborated, strong-consensus, and unanimous are nested agreement views that may provide grounds for multi-witness validation; they are not additional top-level scopes.

At v6.0, 20{,}209 GraphAssertions are exposed in the trusted view, and 7{,}211 remain prediction-only. No deprecated assertion is present because the observed schedule contains no revocation batch. For exclusive reporting, the trusted view is decomposed into 15{,}561 gold-backed GraphAssertions, 4{,}522 validated GraphAssertions in the corroborated non-unanimous component, and 126 validated GraphAssertions in the exclusive non-gold unanimity component; these values sum to 20{,}209. The 5{,}410 ConsensusAssertions are instead the inclusive target-level $C_{6}^{(2)}$ view and can overlap gold-backed material. They must therefore not be added to, or divided by, the configuration-level operational-scope counts.

Across all six GraphVersions, every trusted GraphAssertion has an active qualifying validation ground, no prediction-only GraphAssertion appears in the trusted view, and no GraphAssertion belongs to more than one exclusive operational scope. Thus, validation-record completeness is 100\%, leakage is 0, and scope overlap is 0 for the evaluated states. The gold-scope check confirms correct materialization of corpus annotations; its definitional correctness is a conformance result, not an extractor-accuracy claim.

\subsubsection{Agreement thresholds and gold-aligned correctness}

At fixed graph state v6.0, increasing the setup-support requirement produces the expected selectivity trade-off. Gold-aligned precision rises from 25.3\% for the broad $k\geq1$ exposure view to 90.6\% for full $k=6$ unanimity, while recall decreases from 88.2\% to 16.3\%. The $k\geq1$ value at v6.0 is the union of GraphAssertions whose targets are emitted by any of the six active setups; it must not be confused with the 38.0\% precision of the single S1 setup at v1.0 reported in the cumulative schedule. The inclusive $k=6$ exposure view contains 1{,}236 GraphAssertions; 126 belong to the exclusive non-gold unanimity component after gold-backed material is assigned to the gold scope.

Figure~\ref{fig:precision_vs_setups} and Table~\ref{tab:precision_vs_version} provide the cumulative reading. In the figure, the thick curve shows the macro-averaged precision across TRAM~v2 and AnnoCTR; the thin curves show the per-corpus results; the secondary axis reports macro-averaged recall; and the annotations show the percentage-point change in precision between consecutive states. Each row applies unanimity to the active setup set of that version; the threshold therefore tightens as the number of active setups grows. The two largest precision increases coincide with the introduction of a previously absent generator family, but version, threshold, and composition change together. The table is interpreted as the observed cumulative analytical agreement trajectory, not as an order-independent causal decomposition.

\begin{figure*}[t]
\centering
\includegraphics[width=\textwidth]{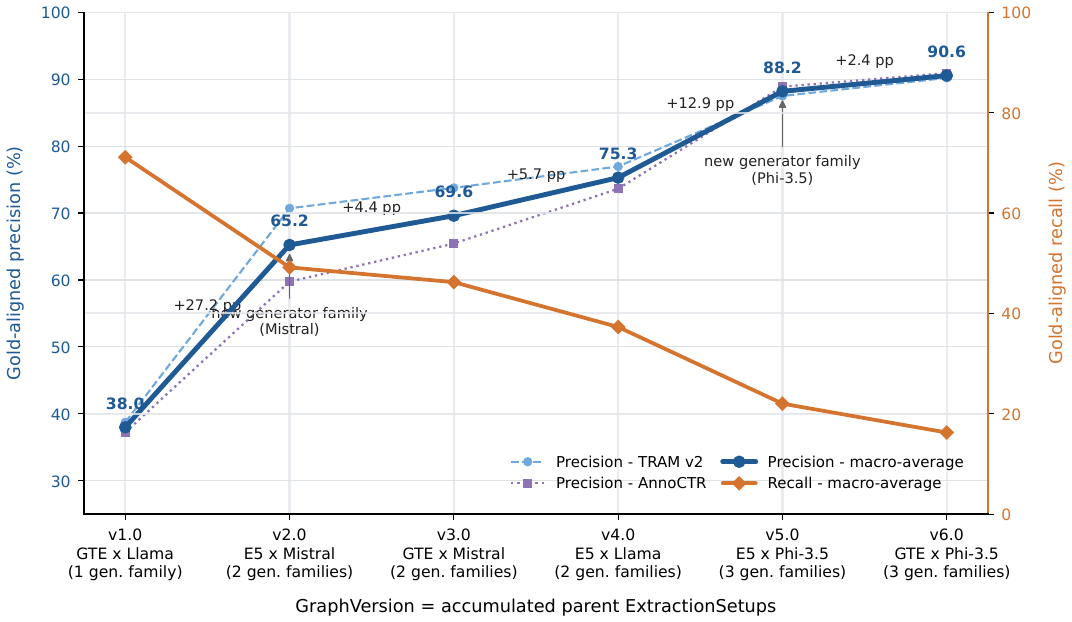}
\caption{Document-level gold-aligned precision and recall of the cumulative unanimity exposure view, v1.0--v6.0.}
\label{fig:precision_vs_setups}
\end{figure*}

\begin{table*}[t]
\caption{Document-level gold-aligned correctness of the cumulative unanimity exposure view, v1.0--v6.0; $^\dagger$ marks the first setup from a new generator family.}
\label{tab:precision_vs_version}
\centering
\small
\begin{tabular}{@{}llcrrcccr@{}}
\toprule
 & & \#Gen. & View & \# ATT\&CK & \multicolumn{3}{c}{\textbf{Gold-aligned precision (\%)}} & $\Delta P$ \\
\cmidrule(lr){6-8}
\textbf{Version} & \textbf{New setup} & \textbf{fam.} & \textbf{size} & \textbf{IDs} & \textbf{TRAM} & \textbf{AnnoCTR} & \textbf{Macro-avg} & \textbf{(pp)} \\
\midrule
v1.0 & GTE$\times$Llama & 1 & 5{,}249 & 120 & 38.7 & 37.2 & \textbf{38.0} & -- \\
v2.0 & E5$\times$Mistral$^\dagger$ & 2 & 3{,}078 & 93 & 70.7 & 59.7 & \textbf{65.2} & $+27.2$ \\
v3.0 & GTE$\times$Mistral & 2 & 3{,}984 & 85 & 73.8 & 65.4 & \textbf{69.6} & $+4.4$ \\
v4.0 & E5$\times$Llama & 2 & 3{,}616 & 68 & 77.0 & 73.6 & \textbf{75.3} & $+5.7$ \\
v5.0 & E5$\times$Phi-3.5$^\dagger$ & 3 & 1{,}615 & 56 & 87.5 & 88.9 & \textbf{88.2} & $+12.9$ \\
v6.0 & GTE$\times$Phi-3.5 & 3 & 1{,}236 & 50 & 90.2 & 90.9 & \textbf{90.6} & $+2.4$ \\
\bottomrule
\end{tabular}
\end{table*}

\paragraph{RQ3 finding}
For all evaluated versions, trusted-view membership is fully explained by active recorded validation grounds, operational scopes remain exclusive, and no prediction-only assertion leaks into the trusted view. Stricter setup-support views yield smaller exposure sets with higher gold-aligned precision. The results therefore support controlled, inspectable promotion rather than silent conversion of Predictions into trusted facts.


\subsection{RQ4: Graph-Native Operational Utility}\label{sec:baseline}\label{sec:baseline-flat}

RQ4 asks which tasks can be answered directly from the governed store. The baseline is a minimal flat extraction output: one record per emitted sentence — technique mapping — without first-class validation history, active-state dependencies, RetrievedContext links, or graph-version semantics. This is a lifecycle baseline, not a claim that relational storage cannot be extended to reproduce the graph schema.

Table~\ref{tab:baseline_flat} evaluates seven operational questions against the information retained by the two representations. A check mark means that the supplied snapshot and audit-query set contain the required entities and relations; a partial mark means that the flat row can return an approximate answer but lacks the governed evidence or lifecycle context.

\begin{table}[t]
\caption{Operational query workload for the flat-output baseline and \tool{}.}
\label{tab:baseline_flat}
\centering
\small
\setlength{\tabcolsep}{3pt}
\begin{tabularx}{\columnwidth}{@{}c>{\raggedright\arraybackslash}Xcc@{}}
\toprule
 & \textbf{Operational question} & \textbf{Flat} & \textbf{KG} \\
\midrule
Q1 & Which native evidence unit supports this ATT\&CK assertion? & \pmark & \cmark \\
Q2 & Which setup, method, run, seed, prompt, and context produced it? & \xmark & \cmark \\
Q3 & What are its operational scope, agreement level, and validation grounds? & \xmark & \cmark \\
Q4 & Which assertions and trust decisions differ across graph versions? & \xmark & \cmark \\
Q5 & Which active assertions and validation grounds depend on a setup selected for revocation? & \xmark & \cmark \\
Q6 & Which disagreements are associated with retriever or generator changes? & \xmark & \cmark \\
Q7 & Which prediction-only assertions satisfy a review-queue policy? & \pmark & \cmark \\
\bottomrule
\end{tabularx}
\end{table}

\paragraph{Evidence and provenance queries}
Q1 follows the GraphAssertion--Prediction--EvidenceUnit path while preserving sentence- versus span-level support. Q2 continues through LLMRun and ExtractionSetup and can also recover prompting method and RetrievedContext. A flat sentence--label row may retain the sentence, hence the partial result for Q1, but it does not contain the complete governed path assumed by Q2.

\paragraph{Trust and version queries}
Q3 retrieves the exclusive operational scope separately from agreement support and validation grounds. Q4 filters by GraphVersion and ImportBatch rather than re-running any extractor. These queries rely on the lifecycle objects introduced in Sections~\ref{sec:trust} and~\ref{sec:evolution}.

\paragraph{Counterfactual revocation}
Q5 traverses from a selected ExtractionSetup to its configurations, Predictions, GraphAssertions, support relations, and policy-derived validation records. It identifies which active targets would lose support and which GraphAssertions would require scope reevaluation before a revocation batch is published. The evaluated schedule does not contain an actual revocation event, so the deprecated scope remains empty; RQ4 demonstrates stored dependency answerability and non-destructive version support, not a measured live-SOC recovery time.

\paragraph{Disagreement and review queries}
Q6 groups disagreement edges by shared or changed retriever and generator components. Q7 selects prediction-only GraphAssertions under an explicit queue policy while retaining evidence and provenance for analyst inspection. A minimal flat output can approximate a queue from confidence or frequency, which motivates the partial mark, but it cannot express the same trust- and provenance-aware selection without enriching the representation or reconstructing missing lifecycle state.

\noindent\textbf{Relation to prior CTI stores}\label{sec:baseline-priorkgs}
The comparison in Section~\ref{sec:related} shows that extraction pipelines, CTI knowledge graphs, sharing platforms, and general provenance models address complementary parts of this lifecycle. The operational contribution evaluated here is not just graph storage, but the joint availability of run-level Predictions, configuration-level GraphAssertions, target-level corroboration, validation records, and versioned active state.

\paragraph{RQ4 finding}
All seven operational questions are directly answerable from the published \tool{} snapshot and audit-query set. A minimal per-emission flat output does not retain sufficient lifecycle state to answer Q2--Q6 directly and provides only approximate answers to Q1 and Q7; answering the same questions would require enriching the representation or reconstructing provenance, validation, and version dependencies through additional processing. The current experiment evaluates answerability rather than analyst usability or query latency.


\subsection{Summary of Findings}\label{sec:evaluation-summary}

\noindent\textbf{RQ1.} Six successive setups are incorporated without schema modification, while complete provenance paths, native evidence handling, scope disjointness, and version-localisability are preserved across all observed graph states.

\noindent\textbf{RQ2.} The same governance model accepts all six heterogeneous setup cells. Cross-generator-family setup pairs exhibit greater behavioral diversity than pairs sharing a generator family, and cumulative full-matrix agreement is associated with higher gold-aligned precision. Together, these analyses provide converging, though not controlled, evidence that heterogeneous corroboration is informative in this instantiation; they do not compare cross-family and within-family precision at a fixed support size.

\noindent\textbf{RQ3.} Every trusted GraphAssertion is backed by an active qualifying validation ground, the four operational scopes remain exclusive, and no prediction-only assertion is exposed as trusted. Increasing the setup-support requirement produces a smaller but more precise agreement-conditioned exposure view.

\noindent\textbf{RQ4.} The graph directly supports evidence tracing, producer reconstruction, trust inspection, version comparison, setup-dependency analysis, disagreement attribution, and review-queue construction. These results establish lifecycle answerability for the stored artifact; they do not constitute a live SOC user study or a performance benchmark.

\section{Discussion}
\label{sec:discussion}

The results support the central claim of this work: fallible TTP-extraction outputs can be governed as auditable CTI assertions without losing evidence, provenance, validation grounds, or version history, and without treating every Prediction as trusted knowledge. The strongest evidence concerns schema-stable ingestion, complete provenance paths, explicit validation grounds, exclusive operational scopes, and lifecycle queryability. The agreement analyses additionally show that stricter setup-support requirements select smaller exposure views with higher gold-aligned precision, at the cost of recall; they do not establish statistical independence or a causal model-family effect.

\subsection{Interpretation of the Research Questions}
\label{sec:rq-answers}

\paragraph{RQ1: Auditable evolution}
Across v1.0--v6.0, all six ExtractionSetups are incorporated through the same schema and ingestion procedure. Every checked GraphAssertion remains traceable to its native evidence unit, supporting Predictions, LLMRuns, assertion configuration, and parent ExtractionSetup; the four operational scopes remain disjoint; and each state is localizable through its ImportBatch and GraphVersion. The relevant result is preservation of existing audit paths under successive ingestion, not graph growth itself. Because the observed schedule is additive, RQ1 does not empirically cover ontology migration, conflicting corpus revisions, or an executed revocation. The methodology nevertheless separates append-only audit history from the version-dependent trusted view, which may contract after revocation or a policy change.

\paragraph{RQ2: Extractor-independent governance and witness diversity}
The structural part of RQ2 is supported: the same representation, canonicalization rules, loader, and trust policy govern all six retriever--generator setups. RAG and RAG+FSP remain distinct assertion configurations.
However, because both configurations inherit the same parent ExtractionSetup, they contribute at most one setup witness; repeated seeds likewise add no witnesses. Here, \emph{extractor-independent} means that governance is not tied to one extractor; it does not imply independent errors.

The empirical conclusion is narrower. Cross-generator-family pairs exhibit greater output-set diversity than pairs sharing a generator family. Separately, unanimity over the progressively larger and more heterogeneous active set is associated with higher gold-aligned precision, reaching 90.6\% at v6.0. These analyses provide converging, though not controlled, evidence that heterogeneous corroboration is informative in this instantiation. They do not compare cross-family and within-family corroboration at fixed support size: threshold, active-set composition, and ingestion order co-vary, while Jaccard distance measures diversity rather than correctness conditional on agreement. No causal family-diversity or statistical-independence claim follows.

\paragraph{RQ3: Controlled promotion to trusted knowledge}
Across the evaluated GraphVersions, every trusted GraphAssertion has an active qualifying validation ground, no prediction-only GraphAssertion enters the trusted view, and the \emph{gold}, \emph{validated}, \emph{prediction-only}, and \emph{deprecated} scopes remain mutually exclusive. Storing a Prediction therefore does not silently promote its GraphAssertion. Stricter setup support yields smaller exposure views with higher gold-aligned precision and lower recall, making the threshold an explicit policy choice. Corroborated, strong-consensus, and unanimous remain nested analytical views rather than additional operational scopes, and a ConsensusAssertion remains corroboration evidence rather than a trusted fact. The evaluation exercises gold and multi-witness grounds, but not analyst validation; it establishes policy conformance, not the effectiveness or cost of human review.

\paragraph{RQ4: Operational utility}
The graph directly supports evidence tracing, producer reconstruction, trust inspection, version comparison, setup-dependency analysis, disagreement attribution, and provenance-aware review queues through the lifecycle it represents, linking Predictions, GraphAssertions, ConsensusAssertions, validation records, and GraphVersions. This is a representational result, not a claim that graph databases are uniquely capable of these operations: an enriched relational or event-sourced implementation could encode the same semantics. The evaluated minimal flat output does not retain them without enrichment or reprocessing. The experiment establishes answerability over the artifact, not query latency, storage efficiency, analyst usability, improved SOC decisions, or recovery time after revocation; the revocation query exposes dependencies, but no revocation batch is executed.

\subsection{Scientific Contribution and Implications}
\label{sec:discussion-implications}

The principal implication is that extraction performance and claim reliability require different units of analysis. A benchmark evaluates whether an ATT\&CK target is recovered; governance must also preserve which run emitted it, which assertion configuration it represents, which parent setups support its canonical target, why its GraphAssertion is trusted, and what could invalidate that decision. The Prediction--GraphAssertion--ConsensusAssertion separation makes these responsibilities explicit. \tool{} therefore complements automated extraction, operates upstream of threat-sharing platforms, and specializes general provenance and knowledge-graph mechanisms for post-extraction CTI governance.

Agreement must consequently be treated as evidence of validation, not as truth. Unanimity cannot guarantee correctness because setups may share models, training data, prompts, retrievers, corpora, or annotation biases. \tool{} records witness identity and family composition so that support can be inspected and reevaluated. The appropriate policy depends on false-positive cost, tolerance for missed techniques, analyst availability, and downstream use; the study does not identify a universal threshold.

The scientific novelty lies not in any individual mechanism, but in their integration into a single explicit claim lifecycle. Prediction-only GraphAssertions remain available in the audit graph for disagreement analysis and future review without entering the trusted view; gold and validated GraphAssertions support more conservative downstream use; and versioned dependencies permit later correction. Uncertain outputs can thus be preserved without becoming indistinguishable from curated CTI facts.

\subsection{Limitations and Threats to Validity}
\label{sec:threats-validity}

\paragraph{Construct validity}
Gold-aligned correctness depends on the quality and granularity of TRAM v2 and AnnoCTR. Alignment is performed at the document level: an exposed assertion counts as gold-aligned when its technique is annotated anywhere in the same report, so sentence- and span-level correctness is not established, and the reported precision and recall are upper bounds with respect to finer-grained matching. Precision and recall are computed over deduplicated document-level tuples, whereas view sizes are reported in GraphAssertions; comparisons with other units or matching policies require caution. Correctness of documents without gold annotations cannot be assessed directly; the reported precision estimates the reliability of the consensus mechanism only where reference annotations are available. Jaccard distance measures output diversity, not error independence, and operational utility is measured as stored answerability rather than analyst or system performance. The graph ontology uses the ATT\&CK v19.1 snapshot retrieved on 22 May 2026. Identifiers absent from that snapshot are retained as marked placeholders; consequently, their absence in the imported bundles must not be automatically interpreted as an extraction error, because some identifiers may derive from deprecated techniques or older gold annotations.

\paragraph{Internal and conclusion validity}
The cumulative ingestion order is not randomized, and support threshold, active-set size, family composition, and version change jointly. The unanimity trajectory is therefore descriptive, not a causal test of generator-family diversity. Deterministic seeds improve reproducibility but do not estimate stochastic variation, while pooling RAG and RAG+FSP in some summaries limits prompting-method-specific conclusions. The use of precomputed outputs and the absence of analyst validation and revocation batches leave live model updates, end-to-end latency, human-review behavior, and deployed rollback unevaluated. Finally, the high precision of the strictest agreement view is accompanied by substantially lower recall and is not a universally optimal operating point.

\paragraph{External validity}
The experiment covers two public English-language CTI corpora, 65 reports, two retrievers, three generator families, and two prompting methods. It does not represent all CTI styles, languages, ATT\&CK domains, proprietary sources, commercial models, or extraction architectures. Corroboration may differ with shared training lineage, common retrieval collections, or additional model families; the reported thresholds are therefore policy examples for this artifact. Neo4j is one realization of the storage-independent methodology; alternative data models, production-scale concurrency, access control, long-term schema migration, and ATT\&CK evolution remain unevaluated.

Overall, the evidence supports \tool{} as a framework for making trust decisions explicit, inspectable, versioned, and subject to non-destructive revocation, not as a mechanism that guarantees model correctness. Its contribution is the transition from a prediction-only representation to an auditable claim lifecycle in which native evidence, provenance, corroboration, validation, and active trust remain distinct and queryable.

\section{Conclusion}
\label{sec:conclusion} 
The broader implication of this work is that automated CTI extraction should not end when a model emits an ATT\&CK mapping. That output begins a governance process in which its evidence, producer, corroboration, validation grounds, and active status must remain explicit and revisable. By providing a stable governance layer between replaceable extractors and downstream CTI systems, \tool{} enables extraction technology to evolve while preserving the provenance and trust history of earlier outputs. 

The next step is to move from artifact-level conformance to operational validation. This requires exercising analyst review and revocation as actual lifecycle events; comparing same-family and cross-family corroboration under controlled support size and ingestion order; studying how trust policies should vary with the operational costs of false positives and missed techniques; and evaluating portability across additional extractors, CTI sources, storage technologies, and SOC workloads. 

More generally, \tool{} provides a basis for treating automatically extracted CTI not as a collection of static facts, but as a set of versioned and contestable claims whose grounds can be inspected, updated, and withdrawn without erasing their history. This shift from prediction storage to explicit claim governance is the principal direction this work opens.
\section*{Ethical considerations}
This work uses public CTI datasets and public open-weight models for offline, non-operational analysis.
We do not instrument a live SOC or collect non-public adversary, analyst, or victim data.
The study evaluates already generated extraction outputs and does not deploy offensive capabilities or act on operational systems.
We therefore identify no significant additional ethical concerns arising from the study.

\section*{CRediT authorship contribution statement}

\noindent\textbf{Federico Valletta:} Conceptualization, Methodology, Software, Formal analysis, Data curation, Writing -- original draft. \textbf{Giacomo Longo:} Conceptualization, Methodology, Writing -- review \& editing.  \textbf{Enrico Russo:} Conceptualization, Validation, Writing -- review \& editing. \textbf{Alessio Merlo:} Conceptualization, Supervision, Writing -- review \& editing.

\section*{Declaration of competing interest}

The authors declare no known competing financial interests or personal relationships that could appear to influence the work reported in this paper.

\section*{Funding}

This research did not receive any specific grant from funding agencies in the public, commercial, or not-for-profit sectors.

\section*{Declaration of generative AI and AI-assisted technologies in the manuscript preparation process}
No technical contributions, implementations, data, or experiments are AI-generated. 

During the preparation of this work, the authors used GitHub Copilot and Grammarly to improve the manuscript's language and assist with its \LaTeX{} formatting. After using such tools, the authors reviewed and edited the content as needed and take full responsibility for the published article.

\section*{Data availability}
\label{sec:data-availability}

All data and code supporting this study are available as supplementary materials to this article and in the public repository \url{https://github.com/federicovalletta/TRACE-CTI-artifacts}.
The artifact archive, \emph{TRACE-CTI\_\allowbreak artifacts\_\allowbreak v6\_\allowbreak 20260728.tar.gz} (SHA-256 {\scriptsize\ttfamily E76E3AD2DCB658D43BDCEC40\allowbreak 6B75E1BC456BDB25\allowbreak 639ADA49CEE0C144AB71C3BC}), contains the KG schema, CSV loaders, Cypher import scripts, audit queries Q1--Q7, the six extraction setups, and the v6.0 KG snapshot; all outputs are deterministic under the published seeds and reproducible from public model weights.

\bibliographystyle{plainnat}
\bibliography{references}

@inproceedings{buchel2025sok,
  title     = {{SoK}: Automated {TTP} Extraction from {CTI} Reports -- Are We There Yet?},
  author    = {B{\"u}chel, Marvin and Paladini, Tommaso and Longari, Stefano and Carminati, Michele and Zanero, Stefano and Binyamini, Hodaya and Engelberg, Gal and Klein, Dan and Guizzardi, Giancarlo and Caselli, Marco and Continella, Andrea and van Steen, Maarten and Peter, Andreas and van Ede, Thijs},
  booktitle = {Proceedings of the 34th {USENIX} Security Symposium},
  pages     = {4621--4641},
  year      = {2025},
  publisher = {USENIX Association}
}

@article{li2024automated,
  title     = {Automated discovery and mapping {ATT\&CK} tactics and techniques for unstructured cyber threat intelligence},
  author    = {Li, Lingzi and Huang, Cheng and Chen, Junren},
  journal   = {Computers \& Security},
  volume    = {140},
  pages     = {103815},
  year      = {2024},
  publisher = {Elsevier},
  doi       = {10.1016/j.cose.2024.103815}
}

@inproceedings{liao2016acing,
  title     = {Acing the {IOC} Game: Toward Automatic Discovery and Analysis of Open-Source Cyber Threat Intelligence},
  author    = {Liao, Xiaojing and Yuan, Kan and Wang, XiaoFeng and Li, Zhou and Xing, Luyi and Beyah, Raheem},
  booktitle = {Proceedings of the 2016 {ACM} {SIGSAC} Conference on Computer and Communications Security ({CCS})},
  pages     = {755--766},
  year      = {2016},
  publisher = {ACM},
  doi       = {10.1145/2976749.2978315}
}

@techreport{strom2020mitre,
  title       = {{MITRE ATT\&CK}: Design and Philosophy},
  author      = {Strom, Blake E. and Applebaum, Andy and Miller, Doug P. and Nickels, Kathryn C. and Pennington, Adam G. and Thomas, Cody B.},
  institution = {The {MITRE} Corporation},
  year        = {2020},
  url         = {https://attack.mitre.org}
}

@misc{ctid2023tram,
  title        = {{TRAM} v2: Threat Report {ATT\&CK} Mapper},
  author       = {{Center for Threat-Informed Defense}},
  howpublished = {\url{https://github.com/center-for-threat-informed-defense/tram}},
  year         = {2023}
}

@inproceedings{lange2024annoctr,
  title     = {{AnnoCTR}: A Dataset for Detecting and Linking Entities, Tactics, and Techniques in Cyber Threat Reports},
  author    = {Lange, Lukas and Reiter, Marc and Str{\"o}tgen, Jannik},
  booktitle = {Proceedings of the 2024 Joint International Conference on Computational Linguistics, Language Resources and Evaluation ({LREC-COLING})},
  pages     = {1052--1065},
  year      = {2024}
}

@inproceedings{rastogi2020malont,
  title     = {{MALOnt}: An Ontology for Malware Threat Intelligence},
  author    = {Rastogi, Nidhi and Dutta, Sharmishtha and Zaki, Mohammed J. and Gittens, Alex and Aggarwal, Charu},
  booktitle = {International Workshop on Deployable Machine Learning for Security Defense ({MLHat})},
  pages     = {28--44},
  year      = {2020},
  publisher = {Springer}
}

@misc{neo4j2024,
  title        = {{Neo4j} Graph Database},
  author       = {{Neo4j, Inc.}},
  howpublished = {\url{https://neo4j.com}},
  year         = {2024}
}

@misc{meta2024llama,
  title        = {The {Llama 3} Herd of Models},
  author       = {{Meta AI}},
  howpublished = {arXiv preprint arXiv:2407.21783},
  year         = {2024}
}

@misc{alibaba2024gte,
  title        = {{GTE-Qwen2-7B-Instruct}: General Text Embeddings},
  author       = {{Alibaba NLP}},
  howpublished = {\url{https://huggingface.co/Alibaba-NLP/gte-Qwen2-7B-instruct}},
  year         = {2024}
}

@misc{wang2022e5,
  title        = {Text Embeddings by Weakly-Supervised Contrastive Pre-training},
  author       = {Wang, Liang and Yang, Nan and Huang, Xiaolong and Jiao, Binxing and Yang, Linjun and Jiang, Daxin and Majumder, Rangan and Wei, Furu},
  howpublished = {arXiv preprint arXiv:2212.03533},
  year         = {2022}
}

@misc{jiang2023mistral,
  title        = {{Mistral 7B}},
  author       = {Jiang, Albert Q. and Sablayrolles, Alexandre and Mensch, Arthur and Bamford, Chris and Chaplot, Devendra Singh and de las Casas, Diego and Bressand, Florian and Lengyel, Gianna and Lample, Guillaume and Saulnier, Lucile and Lavaud, L\'{e}lio Renard and Lachaux, Marie-Anne and Stock, Pierre and Le Scao, Teven and Lavril, Thibaut and Wang, Thomas and Lacroix, Timoth\'{e}e and El Sayed, William},
  howpublished = {arXiv preprint arXiv:2310.06825},
  year         = {2023}
}

@misc{legoy2020rcatt,
  title        = {{rcATT}: A Tool for Automatic Classification of Threat Reports with {MITRE ATT\&CK} Tactics and Techniques},
  author       = {Legoy, Valentine and Caselli, Marco and Seifert, Christian and Peter, Andreas},
  howpublished = {arXiv preprint arXiv:2004.14322},
  year         = {2020}
}

@misc{alam2023looking,
  title        = {Looking Beyond {IoCs}: Automatically Extracting Attack Patterns from External {CTI}},
  author       = {Alam, Md Tanvirul and Bhusal, Dipkamal and Park, Youngja and Rastogi, Nidhi},
  howpublished = {arXiv preprint arXiv:2211.01753},
  year         = {2023}
}

@misc{hemberg2020linking,
  title        = {Linking Threat Tactics, Techniques, and Patterns with Defensive Weaknesses, Vulnerabilities and Affected Platform Configurations for Cyber Hunting},
  author       = {Hemberg, Erik and Kelly, Jonathan and Shlapentokh-Rothman, Michal and Reinstadler, Bryn and Xu, Katherine and Rutar, Nick and O'Reilly, Una-May},
  howpublished = {arXiv preprint arXiv:2010.00533},
  year         = {2020}
}

@inproceedings{wagner2016misp,
  title     = {{MISP}: The Design and Implementation of a Collaborative Threat Intelligence Sharing Platform},
  author    = {Wagner, Cynthia and Dulaunoy, Alexandre and Wagener, G{\'e}rard and Iklody, Andras},
  booktitle = {Proceedings of the 2016 {ACM} Workshop on Information Sharing and Collaborative Security ({WISCS})},
  pages     = {49--56},
  year      = {2016},
  publisher = {ACM}
}

@techreport{lebo2013provo,
  title       = {{PROV-O}: The {PROV} Ontology},
  author      = {Lebo, Timothy and Sahoo, Satya and McGuinness, Deborah and Belhajjame, Khalid and Cheney, James and Corsar, David and Garijo, Daniel and Soiland-Reyes, Stian and Zednik, Stephan and Zhao, Jun},
  institution = {{W3C} Recommendation},
  year        = {2013},
  url         = {https://www.w3.org/TR/prov-o/}
}

@inproceedings{husari2017ttpdrill,
  title     = {{TTPDrill}: Automatic and Accurate Extraction of Threat Actions from Unstructured Text of {CTI} Sources},
  author    = {Husari, Ghaith and Al-Shaer, Ehab and Ahmed, Mohiuddin and Chu, Bill and Niu, Xi},
  booktitle = {Proceedings of the 33rd Annual Computer Security Applications Conference ({ACSAC})},
  pages     = {103--115},
  year      = {2017},
  publisher = {ACM},
  doi       = {10.1145/3134600.3134646}
}

@inproceedings{satvat2021extractor,
  title     = {{EXTRACTOR}: Extracting Attack Behavior from Threat Reports},
  author    = {Satvat, Kiavash and Gjomemo, Rigel and Venkatakrishnan, V. N.},
  booktitle = {Proceedings of the 6th {IEEE} European Symposium on Security and Privacy ({EuroS\&P})},
  pages     = {598--615},
  year      = {2021},
  publisher = {IEEE},
  doi       = {10.1109/EuroSP51992.2021.00046}
}

@inproceedings{li2022attackg,
  title     = {{AttacKG}: Constructing Technique Knowledge Graph from Cyber Threat Intelligence Reports},
  author    = {Li, Zhenyuan and Zeng, Jun and Chen, Yan and Liang, Zhenkai},
  booktitle = {Proceedings of the 27th European Symposium on Research in Computer Security ({ESORICS})},
  pages     = {589--609},
  year      = {2022},
  publisher = {Springer},
  doi       = {10.1007/978-3-031-17140-6_29}
}

@inproceedings{milajerdi2019poirot,
  title     = {{POIROT}: Aligning Attack Behavior with Kernel Audit Records for Cyber Threat Hunting},
  author    = {Milajerdi, Sadegh M. and Eshete, Birhanu and Gjomemo, Rigel and Venkatakrishnan, V. N.},
  booktitle = {Proceedings of the 2019 {ACM} {SIGSAC} Conference on Computer and Communications Security ({CCS})},
  pages     = {1795--1812},
  year      = {2019},
  publisher = {ACM},
  doi       = {10.1145/3319535.3363217}
}

@article{piplai2020creating,
  title   = {Creating Cybersecurity Knowledge Graphs from Malware After Action Reports},
  author  = {Piplai, Aritran and Mittal, Sudip and Joshi, Anupam and Finin, Tim and Holt, James and Zak, Richard},
  journal = {IEEE Access},
  volume  = {8},
  pages   = {211691--211703},
  year    = {2020},
  doi     = {10.1109/ACCESS.2020.3039234}
}

@inproceedings{lewis2020rag,
  title     = {Retrieval-Augmented Generation for Knowledge-Intensive {NLP} Tasks},
  author    = {Lewis, Patrick and Perez, Ethan and Piktus, Aleksandra and Petroni, Fabio and Karpukhin, Vladimir and Goyal, Naman and K{\"u}ttler, Heinrich and Lewis, Mike and Yih, Wen-tau and Rockt{\"a}schel, Tim and Riedel, Sebastian and Kiela, Douwe},
  booktitle = {Advances in Neural Information Processing Systems ({NeurIPS})},
  volume    = {33},
  pages     = {9459--9474},
  year      = {2020}
}

@article{ji2023hallucination,
  title   = {Survey of Hallucination in Natural Language Generation},
  author  = {Ji, Ziwei and Lee, Nayeon and Frieske, Rita and Yu, Tiezheng and Su, Dan and Xu, Yan and Ishii, Etsuko and Bang, Yejin and Madotto, Andrea and Fung, Pascale},
  journal = {ACM Computing Surveys},
  volume  = {55},
  number  = {12},
  pages   = {1--38},
  year    = {2023},
  doi     = {10.1145/3571730}
}

@article{hogan2021kg,
  title   = {Knowledge Graphs},
  author  = {Hogan, Aidan and Blomqvist, Eva and Cochez, Michael and d'Amato, Claudia and de Melo, Gerard and Gutierrez, Claudio and Kirrane, Sabrina and Labra Gayo, Jos{\'e} Emilio and Navigli, Roberto and Neumaier, Sebastian and Ngonga Ngomo, Axel-Cyrille and Polleres, Axel and Rashid, Sabbir M. and Rula, Anisa and Schmelzeisen, Lukas and Sequeda, Juan and Staab, Steffen and Zimmermann, Antoine},
  journal = {ACM Computing Surveys},
  volume  = {54},
  number  = {4},
  pages   = {1--37},
  year    = {2021},
  doi     = {10.1145/3447772}
}

@inproceedings{alahmadi2022falsepositives,
  title     = {99\% False Positives: A Qualitative Study of {SOC} Analysts' Perspectives on Security Alarms},
  author    = {Alahmadi, Bushra A. and Axon, Louise and Martinovic, Ivan},
  booktitle = {Proceedings of the 31st {USENIX} Security Symposium},
  pages     = {2783--2800},
  year      = {2022},
  publisher = {USENIX Association}
}

@inproceedings{orbinato2022automatic,
  title     = {Automatic Mapping of Unstructured Cyber Threat Intelligence: An Experimental Study (Practical Experience Report)},
  author    = {Orbinato, Vittorio and Barbaraci, Mariarosaria and Natella, Roberto and Cotroneo, Domenico},
  booktitle = {Proceedings of the 33rd {IEEE} International Symposium on Software Reliability Engineering ({ISSRE})},
  pages     = {181--192},
  year      = {2022},
  publisher = {IEEE},
  doi       = {10.1109/ISSRE55969.2022.00031}
}

@inproceedings{sundaramurthy2015burnout,
  title     = {A Human Capital Model for Mitigating Security Analyst Burnout},
  author    = {Sundaramurthy, Sathya Chandran and Bardas, Alexandru G. and Case, Jacob and Ou, Xinming and Wesch, Michael and McHugh, John and Rajagopalan, S. Raj},
  booktitle = {Proceedings of the Eleventh Symposium On Usable Privacy and Security ({SOUPS})},
  pages     = {347--359},
  year      = {2015},
  publisher = {USENIX Association}
}

@techreport{oasis2021stix,
  title       = {{STIX} Version 2.1, {OASIS} Standard},
  author      = {{OASIS Open}},
  institution = {OASIS},
  year        = {2021},
  url         = {https://docs.oasis-open.org/cti/stix/v2.1/stix-v2.1.html}
}

@misc{abdin2024phi3,
  title        = {Phi-3 Technical Report: A Highly Capable Language Model Locally on Your Phone},
  author       = {Abdin, Marah and Aneja, Jyoti and Awadalla, Hany and Awadallah, Ahmed and Awan, Ammar Ahmad and Bach, Nguyen and Bahree, Amit and Bakhtiari, Arash and Bao, Jianmin and Behl, Harkirat and others},
  year         = {2024},
  eprint       = {2404.14219},
  archivePrefix= {arXiv},
  primaryClass = {cs.CL}
}

@inproceedings{dietterich2000ensemble,
  title     = {Ensemble Methods in Machine Learning},
  author    = {Dietterich, Thomas G.},
  booktitle = {Multiple Classifier Systems (MCS)},
  series    = {Lecture Notes in Computer Science},
  volume    = {1857},
  pages     = {1--15},
  year      = {2000},
  publisher = {Springer}
}

@article{breiman2001random,
  title   = {Random Forests},
  author  = {Breiman, Leo},
  journal = {Machine Learning},
  volume  = {45},
  number  = {1},
  pages   = {5--32},
  year    = {2001},
  publisher = {Springer},
  doi     = {10.1023/A:1010933404324}
}

\appendix
\appendix
\section{The \tool{} Schema}\label{app:schema}

This appendix provides a compact implementation reference for the
persistent entities that realize the methodological objects introduced in
Section~\ref{sec:primitives} and the graph paths described in
Section~\ref{sec:schema}. Table~\ref{tab:core_entities} preserves the
distinctions used throughout the paper: source evidence is separate from
retrieved context; run-level Predictions are separate from
configuration-level GraphAssertions; and cross-setup ConsensusAssertions
summarize corroboration without becoming trusted facts. Assertion configuration is the logical key formed by an ExtractionSetup and one of the two prompting methods: RAG or RAG+FSP; it is not an additional node class. Likewise, validation and revocation are represented by explicit metadata and relations associated with the governed objects and their GraphVersions rather than requiring separate node labels. The same schema is used in v1.0--v6.0; successive setup ingestion changes graph contents and active state, not the entity model. In the published v6.0 snapshot, Dataset, SourceFile, PromptTemplate, and EvidenceSpan are realized as properties of the materialized nodes or as external source artifacts rather than as separate node labels; all other entities are materialized as nodes.

\begin{table*}[!t]
\setlength{\tabcolsep}{3pt}
\caption{Core persistent entities in the \tool{} schema and their
lifecycle roles.}
\label{tab:core_entities}
\centering
\small
\begin{tabular}{@{}p{2.8cm}p{6.3cm}p{7.2cm}@{}}
\toprule
\textbf{Entity} & \textbf{Representation} & \textbf{Lifecycle role} \\
\midrule
Dataset & A labeled or unlabeled CTI corpus. & Defines an ingestion and
provenance boundary; a dataset is not itself a validation ground unless its
annotations are explicitly designated as trusted. \\
SourceFile
&
A source artifact ingested by the framework, including a CTI report file or one of the MITRE ATT\&CK STIX bundles used by the grounding layer.
&
Preserves file-level origin for reports and ontology sources on the audit path and supports source deactivation or revocation without deleting prior history. \\
Report & One CTI document contained in a Dataset and derived from a
SourceFile. & Provides document-level provenance and review context for its
evidence units. \\
Sentence & A sentence in a Report and the textual unit consumed by the
evaluated extractor. & Preserves sentence-level evidence and contains any
native span-level annotations. \\
EvidenceSpan & A source-identified substring linked to its enclosing
Sentence. & Preserves native span granularity; it is never inferred from a
sentence-level label. \\
\midrule
ExtractionSetup & A versioned retriever--generator bundle with setup-level
parameters. & Identifies the parent witness used for setup-deduplicated
corroboration. Prompting methods and repeated seeds do not create additional
setup witnesses. \\
LLMRun & One execution of an assertion configuration over a dataset and
seed. & Records execution provenance and reproducibility information for
the Predictions emitted in that run. \\
PromptTemplate & The prompt specification used by an LLMRun. & Makes the
prompting method and model input reconstructible without treating the prompt
as source evidence. Records whether the run uses RAG or RAG+FSP and, for the latter, the five few-shot examples included in the prompt. \\
RetrievedContext & A retrieved item supplied to a RAG run. & Records what
the generator was shown; it remains distinct from the native evidence
supporting the ATT\&CK mapping. \\
Prediction & One immutable run-level extractor output. & Records observed
model behavior and is untrusted by default; each Prediction resolves to one
configuration-level GraphAssertion. \\
\midrule
AttackTactic & A MITRE ATT\&CK tactic represented in the grounding layer.
& Organizes normalized techniques by adversary objective. \\
AttackTechnique & A normalized MITRE ATT\&CK technique or sub-technique imported from the current STIX bundles, or a placeholder for an observed identifier absent from those bundles. & Supplies the target used by Predictions, GraphAssertions, and agreement queries. Placeholder nodes retain otherwise ungroundable prediction or gold identifiers and are marked \texttt{observed\_not\_in\_current\_stix\_bundle}. \\
MalontClass & A MALOnt concept used as optional domain enrichment. & Adds
malware-oriented semantic context without changing claim or trust semantics.
\\
\midrule
GraphAssertion & The configuration-level claim
$g=\langle e,a,c\rangle$, where $c=\langle s,m\rangle$ combines the parent
ExtractionSetup and prompting method. & Aggregates repeated run-level
observations for the same normalized key and is the unit assigned to an
operational scope and exposed through the trusted view. \\
ConsensusAssertion & A materialized summary for a canonical target
$u=\langle e,a\rangle$ supported by at least two distinct active parent
ExtractionSetups. & Stores setup-deduplicated support and witness-family
composition. It supplies corroboration evidence but is not automatically
trusted and is not an additional operational scope. \\
Gold annotation instance & A trusted human-supplied corpus annotation
retained separately from model outputs. & May provide a gold-validation
ground for aligned GraphAssertions; it is not counted as a setup witness.
\\
\midrule
ImportBatch & One ingestion or governance batch. & Records when evidence,
extraction outputs, validation grounds, activations, or revocations enter
the audit history. \\
GraphVersion & An identifiable logical graph state reconstructed from the
accumulated history and active-state metadata. & Localizes scope and trust
decisions over time; prior states remain reconstructible while the active
trusted view may expand or contract. \\
\bottomrule
\end{tabular}
\end{table*}

\FloatBarrier
\clearpage


\end{document}